\documentclass[preprint,12pt,authoryear]{elsarticle}

\usepackage{amssymb}
\usepackage{mathtools}
\usepackage{algorithm}
\usepackage{algpseudocode}
\usepackage{caption}
\usepackage{subcaption}
\usepackage{multirow}
\usepackage[flushleft]{threeparttable}
\usepackage{longtable}

\journal{Neurocomputing}

\begin{document}

\begin{frontmatter}



\title{Theoretical research on generative diffusion models: an overview}


\author[inst1]{Melike Nur Yeğin}

\affiliation[inst1]{organization={Computer Engineering Department, Yıldız Technical University},
            city={Istanbul},
            country={Turkey}}

\author[inst1]{Mehmet Fatih Amasyalı}

\begin{abstract}
    Generative diffusion models showed high success in many fields with a powerful theoretical background. They convert the data distribution to noise and remove the noise back to obtain a similar distribution. Many existing reviews focused on the specific application areas without concentrating on the research about the algorithm. Unlike them we investigated the theoretical developments of the generative diffusion models. These approaches mainly divide into two: training-based and sampling-based. Awakening to this allowed us a clear and understandable categorization for the researchers who will make new developments in the future.
\end{abstract}

\begin{highlights}
    \item We briefly explained existing generative models and discussed why we need diffusion models.
    \item We examined the core studies of the diffusion models in a systematical perspective and explained their relationships and missing points.
    \item We categorized the theoretical research of the diffusion models according to the subjects they have focused.
    \item We explained the evaluation metrics of the diffusion models and give the benchmark results on the most familiar data sets.
    \item We discussed the current status of the literature of the diffusion models and showed some future directions.
\end{highlights}


\end{frontmatter}

\section{Introduction}

Generative diffusion models are a family of generative models that slowly convert the score function or approximate lower bound of the data distribution to noise, then take the noise back in the reverse process and obtain a similar data distribution. In addition to its theoretical background, it showed high success in practice and draw attention in the literature.

When we examine the literature of diffusion models, we come across 3 fundamental studies \cite{ho2020denoising, song2019generative, song2020score}. These studies provide a theoretical foundation and practical results, but there are also areas that need improvement. Main problems are the computation cost of the sampling process, the higher log-likelihood values from other models, and suitability for different modalities. There are so many approaches in the literature that improve the algorithm from various perspectives. 

This study is an overview of the theoretical research on generative diffusion models. Many existing reviews have focused on specific application areas such as computer vision \cite{croitoru2022diffusion, ulhaq2022efficient, li2023diffusion}, natural language processing \cite{zou2023diffusion}, medical imaging \cite{kazerouni2022diffusion}, time series analysis \cite{lin2023diffusion}, text-to-image generation \cite{zhang2023text}, text-to-speech synthesis \cite{zhang2023survey}. Some other studies \cite{cao2022survey, yang2022diffusion} examined both improvements and applications. Since they took a broad perspective, they have focused only on popular research and have not investigated many theoretical studies. Another review \cite{chang2023design} investigated the design fundamentals of the diffusion models. This study has focused on the research about network design and addresses only the contributions of some other approaches.

In this study, we categorized the research as training-based and sampling-based approaches. Under these categories we classified the approaches according to the subjects they have focused. The contributions of our study are summarized below:

\begin{itemize}
    \item We briefly explained existing generative models and discussed why we need diffusion models.
    \item We examined the core studies of the diffusion models in a systematical perspective and explained their relationships and missing points.
    \item We categorized the theoretical research of the diffusion models according to the subjects they have focused.
    \item We explained the evaluation metrics of the diffusion models and give the benchmark results on the most familiar data sets.
    \item We discussed the current status of the literature of the diffusion models and showed some future directions.
\end{itemize}

The rest of the study is organized as follows. In Section 2, we briefly mention existing generative models and explain the advantages of the diffusion models. In Section 3, we examine the core studies of the diffusion models in detail. In Section 4, we review the theoretical research of the diffusion models. In Section 5, we explain the evaluation metrics and the results of benchmarks in the most frequently used data sets. Finally, we give ideas for the future studies in the last section.

\section{Generative models}
\begin{figure}
    \centering
    \begin{subfigure}[b]{0.8\textwidth}
    \centering
    \includegraphics[width=\textwidth]{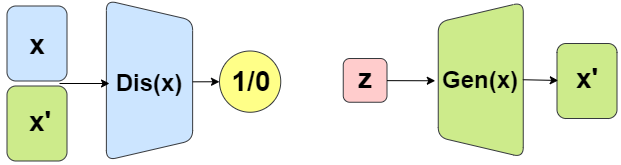}
    \caption{Generative adversarial network}
    \label{fig:gan}
    \end{subfigure}
    
    \begin{subfigure}[b]{0.8\textwidth}
    \centering
    \includegraphics[width=\textwidth]{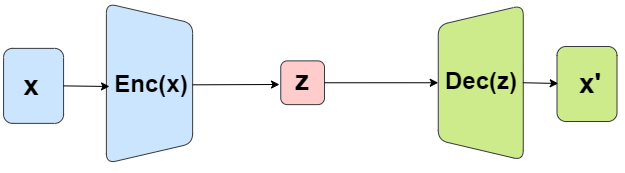}
    \caption{Variational autoencoder}
    \label{fig:vae}
    \end{subfigure}
    
    \begin{subfigure}[b]{0.8\textwidth}
    \centering
    \includegraphics[width=\textwidth]{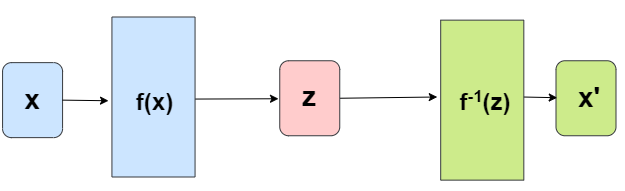}
    \caption{Flow-based model}
    \label{fig:flow}
    \end{subfigure}

    \begin{subfigure}[b]{0.8\textwidth}
    \centering
    \includegraphics[width=\textwidth]{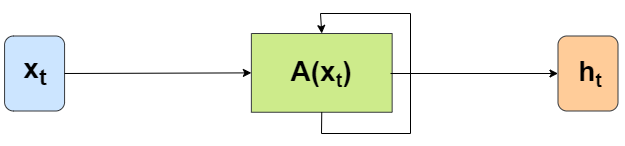}
    \caption{Autoregressive model}
    \label{fig:auto}
    \end{subfigure}

    \begin{subfigure}[b]{0.8\textwidth}
    \centering
    \includegraphics[width=\textwidth]{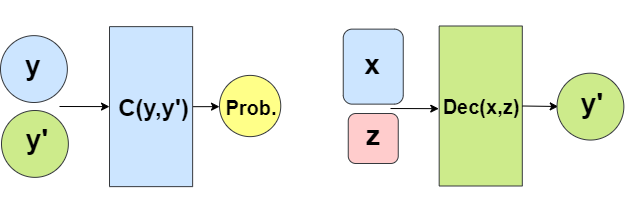}
    \caption{Energy-based model}
    \label{fig:ebm}
    \end{subfigure}

    \begin{subfigure}[b]{0.8\textwidth}
    \centering
    \includegraphics[width=\textwidth]{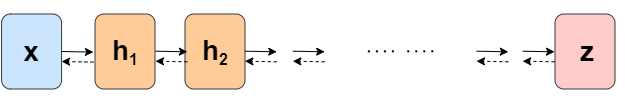}
    \caption{Diffusion model}
    \label{fig:diff}
    \end{subfigure}
    
    \caption{Generative models overview}
    \label{fig:generative}
\end{figure}

Generative models learn the data distribution of the training set with unsupervised learning and then generate new and unknown samples from this distribution. In this section we briefly explain existing 5 main generative models, which are Generative Adversarial Network, Variational Autoencoder, Flow-based model, Autoregressive model, and Energy-based model respectively. After that we introduce generative diffusion models. Figure \ref{fig:generative} shows the structures of the generative models.

Generative adversarial networks (GAN)\cite{goodfellow2020generative} are implicit density models which have two seperate neural networks. First one is discriminator which learns to distinguish real data from fake samples produced by the second one, generator. While these two models are playing the minimax game, adversarial training minimizes the difference between the artificial and real data distributions. That adversarial training causes some disadvantages. First of all, GANs are extremely sensitive to hyperparameter choices. The imbalance between the generator and the discriminator may cause the model parameters become inconsistent, non-convergent or overfit. Another very common situation is mode collapse that the generator produces samples with a limited diversity.

While working with large data sets, the features are numerous and continuous and the calculation of marginal likelihood requires an intractable integral. Variational autoencoders (VAE)\cite{kingma2013auto, rezende2014stochastic} are approximate density models which learn the data distribution by using variational inference (VI) method. The main idea of variational inference is to model the real distribution using a simple distribution e.g. Gaussian, that is easier to evaluate, and to minimize the difference between these two distributions by using the KL-divergence (a metric that measures the difference between distributions). This optimization is called evidence lower bound optimization (ELBO). In VAEs, the encoder network takes the real data to obtain the mean and variance values. Then the data is coded with the reparametrization trick\cite{kingma2013auto} by using these mean and variance values. After that, this coded data is given to the decoder network to obtain an artificial data distribution, and the new data is produced by sampling from this distribution.

In generative models with hidden variables the Gaussian distribution is often used for computational efficiency. But in the real world, most distributions are much more complex than Gaussian. Flow-based models \cite{dinh2014nice} and Normalizing flows \cite{tabak2010density,tabak2013family}, transform a simple distribution into a complex distribution by applying a set of bijective transformation functions in each step of the flow. Normalizing means that a normalized density is obtained after the transformation is applied. The exact log-likehoods of the input data can be calculated and the optimization is performed directly over the negative log-likehood. One of the challenges with normalizing-flow models is the volume of the data. The hidden space becomes very high-dimensional during transformations and this makes interpretation difficult. Also, it is very difficult to implement conditional generation tasks with these models and they produce low-quality samples compared to GAN and VAEs. Also, Gong et al. \cite{gong2021interpreting} shows the hidden connection between normalizing flows and score-based generative models.

Autoregressive models\cite{graves2013generating,van2016pixel}, are tractable density models which works sequentially and predicts the next data with the given previous data. Although they are very successful in the fields such as language modeling and time series forecasting, they only compute polynomial-time probability of the next state and they have difficulty when the next-state probability is hard to compute\cite{lin2021limitations}.

Another family of generative models are energy-based models.(EBM)\cite{lecun2006tutorial,song2021train}. Energy-based models find an unnormalized negative log probability which is called the energy function. Any nonlinear regression function can be chosen as the energy function. The probability distribution is normalized by dividing the energy function to its volume. The flexibility of EBMs provides important benefits in modelling but it is often difficult to calculate and synthesize the exact probability of samples from complex models. Because the parameters of the normalization term typically involve a hard integral.

Generative diffusion models\cite{vincent2011connection,sohl2015deep} is a new family of generative models which eliminates the adversarial training in GANs, the requirement of sequential learning in autoregressive models, the approximate likelihood calculation in VAEs, the volume-growth in normalizing flow models and the sampling difficulty in EBMs. Luo et al. \cite{luo2022understanding} investigated relationship between generative diffusion models and variational autoencoders to understand their underlying logic in a unified perspective. There are two processes in these models; first, the forward process adds noise at multiple scales to the data distribution, gradually converting it to random noise. Second, the reverse process learns a similar of the original data by reversing the diffusion process step by step. 

\section{Core Studies}

There are three main studies about generative diffusion models in the literature. The first study is Denoising Diffusion Probabilistic Models (DDPM)\cite{ho2020denoising} which is inspired from the theory of non-equilibrium thermodynamics. DDPMs use latent variables to estimate the probability distribution. The second core study is Noise Conditional Score Networks (NCSN)\cite{song2019generative}. In NCSNs, a shared network is trained with score matching methods to estimate the score function (gradient of log density) of data distributions at different noise levels. The third study is score-based modeling with stochastic differential equations (Score SDE), \cite{song2020score}, solves the diffusion process with forward and backward SDEs. This study is also important which shows a general framework that puts DDPMs and NCSNs together.  In this section, we will explain these three main studies of diffusion models in detail.

\subsection{Denoising diffusion probabilistic models}

\subsubsection{Definitions \& Notations}
Denoising diffusion probabilistic model\cite{ho2020denoising} is a parameterized Markov chain that is trained using variational inference to produce samples similar to real data after a finite time. There is a two-step process in probabilistic diffusion models: the first is the forward diffusion process and the second is the reverse diffusion or reconstructing process. In the forward diffusion process, we add Gaussian noises repeatedly until the data becomes completely noise. On the other hand, we get the noise back by learning the conditional probability densities with a neural network model in the reverse diffusion process.

Given a data point from a real data distribution $x_0 \sim q(x)$, Equation \ref{eqn:ileri_dif} shows the forward diffusion process that produces a series of noisy samples $x_1,x_2,..x_T$ by adding Gaussian noise step-by-step through the $T$ step. Here, the estimation of the distribution at time $t$ depends only on the distribution just before at time $t-1$. As step $t$ gets larger, sample $x_0$ gradually loses its distinguishable features. When $T \rightarrow \infty$ at the end, $x_T$ is equivalent to an isotropic Gaussian distribution. 

\begin{equation}
    q(x_t | x_{t-1}) = \mathcal{N}(x_t;\sqrt{1-\beta_t}x_{t-1},\beta_tI)
\label{eqn:ileri_dif}
\end{equation}

The whole process is calculated as in Equation \ref{eqn:tum_surec}.

\begin{equation}
    q(x_{1:T} | x_0) = \prod_{t=1}^Tq(x_t | x_{t-1})
    \label{eqn:tum_surec}
\end{equation}

The step sizes indicated by $\beta_t \in (0,1)_{t=1}^T$ in Equation \ref{eqn:ileri_dif} can be taken as constant throughout the process, or can be changed incrementally step by step or parameterized differentially within a certain range. We can reformulate the posterior distribution as in Equation \ref{eqn:reformulate} if we take $\alpha_t=1-\beta_t$ and $\overline{\alpha}_t=\prod_{i=1}^T\alpha_i$. Here $z_{t-1},z_{t-2},..,\sim \mathcal{N}(0,I)$ and $\overline{z}_{t-2}$ is a combination of two Gaussians. For two Gaussians with different variances, (e.g. $\mathcal{N}(0, \sigma_1^2I)$ and $\mathcal{N}(0,\sigma_2^2I)$) their combination is calculated as $\mathcal{N}(0,(\sigma_1^2+\sigma_2^2)I)$, and standart deviation is $\sqrt{(1-\alpha_t)+\alpha_t(1-\alpha_{t-1})}=\sqrt{1-\alpha_t\alpha_{t-1}}$. Usually, we can take a larger update step when the sample becomes noisy $\beta_1<\beta_2,...,\beta_T$, therefore $\overline{\alpha}_1>\overline{\alpha}_2,...,\overline{\alpha}_T$.

\begin{equation}
    \begin{aligned}
        x_t &= \sqrt{\alpha_t}x_{t-1} + \sqrt{1-\alpha_t} z_{t-1}\\
        &=\sqrt{\alpha_t\alpha_{t-1}}x_{t-2} + \sqrt{1-\alpha_t\alpha_{t-1}}\overline{z}_{t-2}\\
        &=...\\
        &=\sqrt{\overline{a}_t}x_0 + \sqrt{1-\overline{a}_t}z\\
        q(x_t|x_0)&=\mathcal{N}(x_t;\sqrt{\overline{a}_t}x_0, (1-\overline{a}_t)I)
    \end{aligned}
    \label{eqn:reformulate}
\end{equation}

For the reverse process, we want to estimate the probability density in the previous step. In order to make an estimation of the previous state, we need to know all previous gradients, and this can only be obtained with a learning model. A neural network model with parameters $\theta$ given in Equation \ref{eqn:ters_dif} can predict the $p_\theta(x_{t-1}|x_t)$. The reverse diffusion process begins with the standard Gaussian distribution $p_\theta(x_T)=\mathcal{N}(x_T;0,I)$.

\begin{equation}
    p_\theta(x_{t-1}|x_t)=\mathcal{N}(x_{t-1};\mu_\theta(x_t,t),\Sigma_\theta(x_t,t))
    \label{eqn:ters_dif}
\end{equation}

The entire process of the reverse diffusion is given in Equation \ref{eqn:tum_ters}.

\begin{equation}
p_\theta(x_{0:T})=p(x_T)\prod_{t=1}^Tp_\theta(x_{t-1}|x_t)
\label{eqn:tum_ters}
\end{equation}

\subsubsection{Training objective}

Denoising diffusion probabilistic models are trained similarly to VAEs. The size of the input layer is the same as the data. The middle layers are linear layers with corresponding activation functions. The output layer has the same size with the input layer, so the original data is reconstructed. The objective of the network model is to optimize the loss function in Equation \ref{eqn:objective}.

\begin{equation}
    \begin{aligned}
        E[-logp_\theta(x_0)] &\leq E_q\Big[-log\frac{p_\theta(x_{0:T})}{q(x_{1:T} | x_0)}\Big]\\
        &=E_q\Big[-logp_\theta(x_T)-\sum_{t\geq1}^Tlog\frac{p_\theta(x_{t-1}|x_t)}{q(x_t|x_{t-1})}\Big]
    \end{aligned}
\label{eqn:objective}
\end{equation}

The loss function gets better by reducing the variance, and it is rewritten in Equation \ref{eqn:loss}. Here $L_T$ is the forward loss showing the difference between the distribution of random noise in the last step of the forward process and the distribution before the last step. This value is a constant that depends on the variance schedule. $L_{1:T-1}$ is the error that represents the sum of the differences between the forward and backward step distributions at each step in the reverse process. $L_0$ is the decoding error.

\begin{equation}
    \begin{aligned}
        \mathcal{L} &= E_q\Big[\underbrace{ D_{KL}(q(x_T|x_0)||p(x_T)}_{L_T}\\ 
        &+ \sum_{t>1}\underbrace{D_{KL}(q(x_{t-1}|x_t,x_0)||p_\theta(x_{t-1}|x_t))}_{L_{t-1}}\\
        &\underbrace{-logp_\theta(x_0|x_1))}_{L_0}\Big]
    \end{aligned}
\label{eqn:loss}
\end{equation}

The only term we can train is $L_{1:{T-1}}$ while minimizing Equation \ref{eqn:loss} with SGD. If we reparametrize the posterior distribution $q(x_{t-1} | x_t, x_0)$ with Bayes rule we obtain Equation \ref{eqn:posterior}.

\begin{equation}
    q(x_{t-1} | x_t, x_0)=\mathcal{N}(x_{t-1}; \overline{\mu}_t(x_t,x_0),\overline{\beta}_tI)
    \label{eqn:posterior}
\end{equation}

The mean and variance here can be calculated as in Equation \ref{eqn:meanvariance}.

\begin{equation}
\begin{aligned}
    \overline{\mu}_t(x_t,x_0)&:=\frac{\sqrt{\alpha_{t-1}}\beta_t}{1-\overline{\alpha}_t}x_0 + \frac{ \sqrt{\alpha_t}(1-\overline{\alpha}_{t-1})}{1-\overline{\alpha}_t}x_t \\
    \overline{\beta}_t&:=\frac{1-\overline{\alpha}_{t-1}}{1-\overline{\alpha}_t}\beta_t
\end{aligned}    
\label{eqn:meanvariance}
\end{equation}

For the calculated distribution in the reverse process $p_\theta(x_{t-1}|x_t)=\mathcal{N}(x_{t-1};\mu_\theta(x_t,t),\Sigma_\theta(x_t,t))$, variance function $\Sigma_\theta(x_t,t)=\sigma_t^2I$ both $\sigma_t^2=\beta_t$ and $\sigma_t^2=\overline{\beta}_t=\frac{1-\overline{\alpha}_{t-1}}{1-\overline{\alpha}_t}\beta_t$ they experimentally obtained the same results. They also used a special parametrization for the mean function $\mu_\theta(x_t,t)$. Write $L_{t-1}$ term in Equation \ref{eqn:loss} as $p_\theta(x_{t-1}|x_t)=\mathcal{N}(x_{t-1};\mu_\theta(x_t,t),\sigma_t^2I)$, $L_{t-1}$ given in Equation \ref{eqn:l2_loss} can be seen as the L2-error between two mean coefficients.

\begin{equation}
    L_{t-1}=E_q\Big[\frac{1}{2\sigma_t^2}||\overline{\mu}_t(x_t,x_0)-\mu_\theta(x_t,t)||^2\Big]+C
    \label{eqn:l2_loss}
\end{equation}

Here $C$ is a constant that does not depend on $\theta$.The best parameterization for $\mu_\theta$ is a model that can predict $\overline{\mu}_t$. For $\epsilon=\mathcal{N}(0,I)$ Equation \ref{eqn:reformulate}, is reparameterized as $x_t(x_0,\epsilon)=\sqrt{\overline{a}_t}x_0+(1-\overline{a}_t)\epsilon)$  and written as in Equation \ref{eqn:loss_formula}.

\begin{equation}
\begin{aligned}
     L_{t-1} - C &= E_{x_0,\epsilon}\Big[\frac{1}{2\sigma_t^2}\Big|\Big|\overline{\mu}_t\Big(x_t(x_0,\epsilon), \frac{1}{\sqrt{\overline{\alpha}_t}}x_t(x_0,\epsilon - \sqrt{1-\overline{\alpha}_t}\epsilon)\Big)-\mu_\theta(x_t(x_0,\epsilon),t)\Big|\Big|^2\Big] \\
     &= E_{x_0,\epsilon}\Big[\frac{1}{2\sigma_t^2}\Big|\Big|\frac{1}{\sqrt{\alpha}_t}\Big(x_t(x_0,\epsilon)- \frac{\beta_t}{\sqrt{1-\overline{a}_t}}\epsilon\Big) -\mu_\theta(x_t(x_0,\epsilon),t)\Big|\Big|^2\Big]
\end{aligned}
\label{eqn:loss_formula}
\end{equation}

Considering that $x_t$ is an input to the model, the parameterization for $\mu_\theta$ can be selected as in Equation \ref{eqn:mean}. Here $\epsilon_\theta$ is a function approximator that aims to estimate $\epsilon$ from $x_t$.

\begin{equation}
    \mu_\theta(x_t,t)=\frac{1}{\sqrt{\alpha_t}}\big(x_t-\frac{\beta_t}{\sqrt{1-\overline{\alpha}_t}}\epsilon_\theta(x_t,t)\big)
\label{eqn:mean}
\end{equation}

Finally, the simplified loss function becomes in Equation \ref{eqn:son_loss}.

\begin{equation}
    L_{t-1}-C=E_{x_0,\epsilon}\Big[\frac{\beta_t^2}{2\sigma_t^2\alpha_t(1-\overline{\alpha}_t)}||\epsilon-\epsilon_\theta(\sqrt{\overline{\alpha}_t}x_0 + \sqrt{1-\overline{\alpha}}_t\epsilon,t)||^2\Big]
\label{eqn:son_loss}
\end{equation}

Ho et al. found that Equation \ref{eqn:simple_loss} variant of the variational bound is more beneficial for sample quality and simpler to implement.

\begin{equation}
    L_{\text{simple}}(\theta):=E_{t,x_0,\epsilon}\Big[||\epsilon-\epsilon_\theta(\sqrt{\overline{\alpha}_t}x_0 + \sqrt{1-\overline{\alpha}}_t\epsilon,t)||^2\Big]
    \label{eqn:simple_loss}
\end{equation}

In short, the mean function can be trained to estimate $\mu_t$ or $\epsilon$ by changing the parameterization of $\mu_\theta$. Estimating $\epsilon$ is similar with Langevin dynamics and simplifies the variational bound of the diffusion model by simulating denoising score matching. 

\subsubsection{Sampling algorithm}

The sampling process is similar to a Langevin dynamics process that learns the gradient of data density through $\epsilon_\theta$. The whole process of sampling is given in Algorithm \ref{alg:ddpm_sampling}. In case $t=T$ here, the process is started by taking $x_T\sim\mathcal{N}(0,I)$. For $z\sim\mathcal{N}(0,I)$ sample $x_{t-1}\sim p_\theta(x_{t-1}|x_t)$ at time $t-1$ is drawn from the distribution with a mean function as in Equation \ref{eqn:mean}. 

\begin{algorithm}
\caption{DDPM sampling algorithm}\label{alg:ddpm_sampling}
\begin{algorithmic}
\State $x_T \sim \mathcal{N}(0,I)$
\For {$t=T,...,1$}
    \If {$t>1$}
        \State $z \sim \mathcal{N}(0,I)$ 
    \Else
        \State $z=0$
    \EndIf
    \State $x_{t-1}=\frac{1}{\sqrt{\alpha_t}}\Big(x_t-\frac{1-\alpha_t}{\sqrt{1-\overline{\alpha}_t}}\epsilon_\theta(x_t,t)\Big) + \sigma_tz$
\EndFor\\
\Return $x_0$
\end{algorithmic}
\end{algorithm}

\subsection{Noise conditional score networks}

\subsubsection{Definitions \& Notations}

Noise conditional score networks\cite{song2019generative} model gradients of the data distribution instead of the probability density. In energy-based models, the probability distribution is expressed as $p_\theta(x)$ in Equation \ref{eqn:dagilim}. Here $E_\theta$ is the energy function. While data points with high probability have low energy, data points with low probability have high energy. Therefore $E_\theta$ has a negative sign. When this function is a neural network, the parameters are denoted by $\theta$ and the input data is represented by $x$. The output of this network is a scalar value between $-\infty$ and $\infty$. The exponential operation ensures that any possible input is assigned with a probability greater than zero. $z_\theta$ in Equation \ref{eqn:dagilim} is the normalization constant and expressed with $\int_xe^{-E_\theta(x)}dx$ when $x$ is continuous. $z_\theta$ ensures that the sum of the density is 1. ($\int_x p_\theta(x)dx=1$)

\begin{equation}
    p_\theta(x)=\frac{e^{-E_\theta(x)}}{z_\theta}
\label{eqn:dagilim}
\end{equation}

To calculate $p_\theta(x)$ in the Equation \ref{eqn:dagilim}, we need to find the normalization constant $z_\theta$. This is an intractable value for large neural networks where the inputs are high dimensional. Energy-based models are trained by some methods such as contrastive divergence\cite{qiu2019unbiased}. We can avoid from difficult calculation of the normalization constant by modeling the score function instead of the probability density.

The score function $s_\theta(x)$ is given in Equation \ref{eqn:skor}. Since the gradient of the normalization constant is $\nabla_xlogz_\theta = 0$, the score function is independent of $z_\theta$. In this way, it is not necessary to use any special architecture to make the normalization constant tractable.

\begin{equation}
\begin{split}
    s_\theta(x) \approx \nabla_xlogp_\theta(x) \\
    s_\theta(x) = \nabla_xE_\theta(x)-\nabla_x log z_\theta \\
    s_\theta(x) = -\nabla_x E_\theta(x)
\end{split}
\label{eqn:skor}
\end{equation}

After the normalization constant problem is eliminated, $p_\theta(x)$ is found as a normalized probability density function and can be trained by maximizing likelihood in Equation \ref{eqn:optim}.

\begin{equation}
    \max_{\theta}\sum\limits_{i=1}^{N}logp_\theta(x_i)
\label{eqn:optim}
\end{equation}

We explained the basics of score-based models so far. However, the sample quality is not satisfying in practice and this problem has been solved by using the diffusion process. When sampling with Langevin dynamics the first sample is most likely from the low-density region when the data is in high-dimensional space. A solution is to add multiple noises to the distribution with a diffusion process. These multiple noises gradually converts the data to random noise. The data distribution is corrupted by different levels of Gaussian noise and the scores of these corrupted distributions are estimated. Adding large noises makes the data significantly different from the original. On the contrary, adding small noises may not completely cover low density areas. Multiple noise scales are used in an order to be sure about the size of the noise. Low-density data regions are filled when the noise is large enough and the predicted scores improved. Assuming that all noises are isotropic Gaussian and their standard deviations are given as $\sigma_1<\sigma_2<..<\sigma_L$, the noisy data distribution is shown as $p_{\sigma_i}(x)$ in Equation \ref{eqn:perturbed}.

\begin{equation}
    p_{\sigma_i}(x) = \int p(y)\mathcal{N}(x;y,\sigma_i^2I)dy
\label{eqn:perturbed}
\end{equation}

Adding noise at multiple scales is critical to the success of score-based generative models. Generalizing the number of noise scales to infinity allows high-quality and controllable generation as well as enabling full log-likelihood calculation. 

\subsubsection{Training objective}
Noise conditional score networks are trained by minimizing the Fisher divergence between the data distribution and the model. The Fisher divergence calculates the square of the $l_2$ distance between the real data score and the score-based model. The optimization term is given in Equation \ref{eqn:fisher}.

\begin{equation}
    E_{p(x)}[||\nabla_x logp(x)-s_\theta(x)||_2^2]
    \label{eqn:fisher}
\end{equation}

It is not possible to calculate this divergence directly because the score $\nabla_xlogp(x)$ of the data distribution is unknown. The score matching networks try to predict the data score by mixing the data at different noise levels. Score-matching networks are optimized directly on a dataset by stochastic gradient descent. The input and output dimensions should be equal as the only architectural requirement for noise conditional score networks.

When the optimization term given in Equation \ref{eqn:fisher} is written as in Equation \ref{eqn:fisher_l2}, the distance $l_2$ is weighted by the coefficient $p(x)$. Therefore, the predicted score functions work incorrectly in low-density regions. 

\begin{equation}
    E_{p(x)}[||\nabla_x logp(x)-s_\theta(x)||_2^2] = \int p(x)||\nabla_x logp(x)-s_\theta(x)||_2^2 dx
\label{eqn:fisher_l2}
\end{equation}

Due to the sampling problem in low-density regions, the Denoising Score Matching (DSM) \cite{vincent2011connection} method given in Equation \ref{eqn:DSM} proposes an effective solution. In this method, the original score density is distorted with an increasing noise sequence. Here $p_\sigma(\overline{x}|x)$ denotes the distorted data distribution. This method is only accurate for noises small enough to be $p_\sigma(\overline{x}|x) \approx p_{data(x)}$.

\begin{equation}
    \frac{1}{2}E_{p_\sigma(\overline{x}|x)p_{data}(x)}[||s_{\theta}(\overline{x})-\nabla_{\overline{x}}logp_\sigma(\overline{x}|x)||_2^2]
    \label{eqn:DSM}
\end{equation}

Here, if distorted data distribution is $p_\sigma(\overline{x}|x) = \mathcal{N}(\overline{x}|x, \sigma^2I)$, it is found as $\nabla_{\overline{x}}logp_\sigma(\overline{x}|x) = -\frac{\overline{x}-x}{\sigma^2}$. Accordingly, for a given value of $\sigma$, the denoising score matching objective is expressed as in Equation \ref{eqn:DSM_final}.

\begin{equation}
    \ell(\theta;\sigma)=\frac{1}{2}E_{p_{data}(x)}E_{\overline{x}\sim\mathcal{N}(x, \sigma^2I)}\Big[\Big|\Big|s_{\theta}(\overline{x})+\frac{\overline{x}-x}{\sigma^2}\Big|\Big|_2^2\Big]
    \label{eqn:DSM_final}
\end{equation}

If Equation \ref{eqn:DSM_final} is combined for all available $\sigma \in \{\sigma_i\}_{i=1}^L$, then the weighted sum of Fisher divergences at different noise scales is given in Equation \ref{eqn:unified_DSM}. Here $\lambda(\sigma_i) \in  R_{>0}$ is the positive weighting function and it is usually chosen as $\lambda(\sigma_i)=\sigma_i^2$ when optimizing score-matching.

\begin{equation}
    \mathcal{L}(\theta;\{\sigma_i\}_{i=1}^L)=\frac{1}{L}\sum_{i=1}^{L}\lambda(\sigma_i)\ell(\theta;\sigma_i)
    \label{eqn:unified_DSM}
\end{equation}

Sliced Score Matching (SSM) \cite{song2020sliced} is another score-matching method that estimates the undistorted real score through forward mode auto-differentiation by projecting the score into a random vector. This method achieves similar results to DSM but requires 4 times more computational cost.

\subsubsection{Sampling algorithm}

Langevin dynamics combined with stochastic gradient descent can generate samples from the distribution using gradients in the Markov chain. Compared to standard SGD, stochastic gradient Langevin dynamics adds Gaussian noise to parameter updates to avoid from local minima. The entire process of sampling is given in the Algorithm \ref{alg:ncsn_sampling}.

It starts by taking a sample $\overline{x}_0$ from a prior distribution of unstructured random noise. These samples and the step size $a_1$ are used to draw samples from $p_{\sigma_1}$. Then, the next step continues with the samples found. The step size gets smaller at each step. Samples are modified in this way to obtain realistic samples.

\begin{algorithm}
\caption{NCSN sampling algorithm}\label{alg:ncsn_sampling}
\begin{algorithmic}
\Require $\{\sigma_i\},\epsilon,T$
\State Initialize $\overline{x}_0$
\For {$i=1,...,L$}
    \State $a_i=\epsilon \cdot \sigma_i^2/\sigma_L^2$
    \For {$t=1,...,T$}
        \State Draw $z_t \sim \mathcal{N}(0, I)$
        \State $\overline{x}_t=\overline{x}_{t-1}+\frac{a_i}{2}s_\theta(\overline{x}_{t-1},\sigma_i)) + \sqrt{a_i}z_t$
    \EndFor
    \State $\overline{x}_0 \gets \overline{x}_T$   
\EndFor\\
\Return $\overline{x}_T$
\end{algorithmic}
\end{algorithm}

Since the noise gradually decreases, this process is called Annealed Langevin dynamics (ALD). In this way, the problem in low-density regions is eliminated. When $\sigma_1$ is chosen large enough, low density regions in $p_{\sigma_1}$ are considerably reduced and high quality samples can be drawn. These samples are a good starting point for the next step, and  provides a high quality sampling from the distribution in the last step $p_{\sigma_L}$.

\subsection{Score-based modeling with Stochastic Differential Equations}

\subsubsection{Definitions \& Notations}
Song and Ermon \cite{song2020score} generalized the noising process to an infinite number of steps in the DDPM and NCSN models and defined it as a continuous-time stochastic process. Stochastic processes are solutions of stochastic differential equations (SDE). SDEs are expressed as in Equation \ref{eqn:sde}. Here $f(.,t): R^d \rightarrow R^d$ is a vector-valued function called the drift coefficient, $g(t) \in R$ is a real-valued function called the diffusion coefficient, $\textbf{w}$ is the symbol of standard Brownian motion and $d\textbf{w}$ is the infinitesimal white noise.

\begin{equation}
    d\textbf{x}=\textbf{f}(\textbf{x},t)dt + g(t)d\textbf{w}
    \label{eqn:sde}
\end{equation}

The choice of SDE is not same always like that there are hundreds of ways to distort the data distribution. Noising processes in DDPMs and NCSNs are different versions of SDEs. The relevant SDEs are given in Equation \ref{eqn:DDPM_SDE} for DDPM and Equation \ref{eqn:NCSN_SDE} for NCSN. Here $\beta(\frac{t}{T})=T\beta_t$ and $\sigma(\frac{t}{T})=\sigma_t$ as T goes to infinity.

\begin{equation}
    d\textbf{x}=-\frac{1}{2}\beta(t)\textbf{x}dt + \sqrt{\beta(t)}d\textbf{w}
    \label{eqn:DDPM_SDE}
\end{equation}

\begin{equation}
    d\textbf{x}=\sqrt{\frac{d[\sigma(t)^2]}{dt}}d\textbf{w}
    \label{eqn:NCSN_SDE}
\end{equation}

Solutions of the reverse-time SDEs are reverse diffusion processes that gradually convert noise into data. There is a reverse-SDE for every SDE as given in Equation \ref{eqn:reverse-sde}. Here, the sample distribution at any time $t$ is denoted by $p_t(x)$. Equation \ref{eqn:sde_denoising} uses denoising score matching method to find a weighted combination of Fisher divergences in reverse time. While this method is computationally more efficient, it can be optimized with the same performance as methods such as sliced score matching \cite{song2020score}. $dt$ denotes infinitesimal time steps in the negative direction, as SDE must be solved in reverse time from $t=T$ to $t=0$. 

\begin{equation}
    d\textbf{x}=[\textbf{f}(\textbf{x},t)-g^2(t)\nabla_xlogp_t(\textbf{x})]dt + g(t)d\textbf{w}
    \label{eqn:reverse-sde}
\end{equation}

For all stochastic diffusion processes, there is a deterministic process corresponds to an ordinary differential equation (ODE). The probability flow ODE given in the Equation \ref{eqn:flow_ode} is an ordinary differential equation whose trajectories have the same marginal with the reverse-time SDE. Both reverse-time SDE and probability flow ODE allow sampling from the same data distribution. Equation \ref{eqn:flow_ode} is an example of a neural ODE, as the score function is predicted by a time-dependent score-based model (a neural network).

\begin{equation}
    d\textbf{x}=\Big[\textbf{f}(\textbf{x},t)-\frac{1}{2}g^2(t)\nabla_xlogp_t(\textbf{x})\Big]dt
    \label{eqn:flow_ode}
\end{equation}

\subsubsection{Training objective}
Stochastic differential equations are solved with a collection of continuous random variables $\{x(t)\}_{t\in[0,T]}$. These random variables follow stochastic trajectories as the time index $t$ goes from the starting time $0$ to the ending time $T$. Suppose the marginal probability density function of $x(t)$ is denoted by $p_t(x)$. Here $t\in[0,T]$ is similar to $i=1,2,...,L$ when we have a finite number of noise and $p_t(x)$ here is similar to $p_{\sigma_i}(x)$. The distribution before any noise is applied at time $t=0$ is the real data distribution $p_0(x)=p(x)$. Adding noise to $p(x)$ by the stochastic process for a sufficiently long time $T$ turns it into an explainable noise distribution $\pi(x)$ which is called the prior distribution. When we add the last and largest noise $\sigma_L$ we get the distribution $p_{\sigma_L}(x)$ and it is similar to $p_T(x)$ when we have finite-scale noise. The SDE given in Equation \ref{eqn:sde} is manually determined, just like $\sigma_1<\sigma_2<...<\sigma_L$ when we have finite scale noise.

A time-dependent score-based model $s_\theta(x,t) \approx \nabla_xlogp_t(\textbf{x})$ is trained to solve the reverse-SDE with score-matching. This model is similar to $s_\theta(x,\sigma_i)$ when we have finite-scale noise. The training objective given in the Equation \ref{eqn:sde_fisher} is a continuous and weighted combination of the Fisher divergences between the score of the real data distribution and the model. Here, $U(0,T)$ denotes a uniform distribution in the time interval $[0,T]$ and $\lambda:R \rightarrow R_{>0}$ is the positive weighting function. It is usually $\lambda(t) \propto 1/E[||\nabla_{x}logp(x(t)|x(0))||_2^2]$ to balance the magnitude of different score-matching losses over time.

\begin{equation}
    E_{t \in U(0,T)}[E_{p_t(x)}\lambda(t)||\nabla_xlogp_t(\textbf{x}) - s_\theta(x,t)||_2^2]
    \label{eqn:sde_fisher}
\end{equation}

\begin{equation}
    \theta^\star=\underset{\theta}{\arg\min} E_t\Big\{\lambda(t)E_{x(0)}E_{x(t)|x(0)}[||s_\theta(x(t),t)-\nabla_{x(t)}logp_{0t}(x(t)|x(0))||^2_2]\Big\}
    \label{eqn:sde_denoising}
\end{equation}

It is necessary to know the transition kernel $p_{0t}(x(t)|x(0))$ to solve the equation \ref{eqn:sde_denoising}. The transition kernel is a Gaussian distribution when $f(.,t)$ is affine and the mean and variance can be obtained usually in closed form with the standard techniques.

We have mentioned that the noise processes used in DDPMs and NCSNs correspond to different versions of SDEs. SDE in Equation \ref{eqn:NCSN_SDE} is always an exploding variance process when $t\rightarrow\infty$. Equation \ref{eqn:DDPM_SDE} has a fixed variance when $t\rightarrow\infty$ and the initial distribution has unit variance. Because of this difference, Equation \ref{eqn:NCSN_SDE} is called variance-exploding(VE) SDE and Equation \ref{eqn:DDPM_SDE} is called variance-protected(VP) SDE. In addition, there is another SDE that works well in likelihood which is called sub-VP SDE.

Since VE, VP, and sub-VP SDEs all have an affine drift coefficient, the transition kernels $p_{0t}(x(t)|x(0))$ are Gaussian and can be calculated in closed form.

\subsubsection{Sampling algorithm}

After training the time-dependent score-based model $s_\theta$, sampling from $p_0$ requires solving the reverse-time SDE with numerical approaches. There are many general purpose numerical solvers that solve SDEs. For example, Euler-Maruyama solver and stochastic Runge-Kutta methods can be applied to reverse-time SDEs to generate samples. The sampling method of DDPM corresponds to a special discretization of the reverse-time VP-SDE. Also, reverse diffusion samplers \cite{song2020score} perform slightly better than classical samplers, which discretize reverse-time SDE in the same way as forward discretization and therefore it is easy to sample when forward discretization is given.

In Predictor-Corrector(PC) samplers, first the numerical SDE solver which is called predictor makes a prediction for the sample in the next step at each time step. Then the score-based MCMC approach which is called corrector corrects the marginal distribution of the sample. PC samplers generalize the original sampling methods of NCSNs and DDPMs. NCSN uses the identity function as the predictor and the annealed Langevin dynamics as the corrector, while the DDPM uses the numerical solvers as the predictor and identity function as the corrector. Algorithms for Predictor-Corrector samplers for VE-SDE and VP-SDE are given in \ref{alg:vesde-pc} and \ref{alg:vpsde-pc}. $N$ specifies the discretization number for reverse-time SDE, and $M$ specifies the step number of the corrector in algorithms. The reverse diffusion SDE solver is used as the predictor method and the annealed Langevin dynamics is used as the corrector method. The step size of the Langevin dynamics is shown as $\{\epsilon_i\}_{i=0}^{N-1}$.

\begin{algorithm}
\caption{VE-SDE predictor-corrector sampling algorithm}\label{alg:vesde-pc}
\begin{algorithmic}
\State $x_N\sim\mathcal{N}(0,\sigma_{max}^2I)$
\For {$i=N-1,...,0$}
    \State $x_i'\leftarrow x_{i+1}+(\sigma_{i+1}^2-\sigma_i^2)s_{\theta^\star}(x_{i+1},\sigma_{i+1})$
    \State $z \sim \mathcal{N}(0, I)$
    \State $x_i\leftarrow x_i'+\sqrt{\sigma_{i+1}^2-\sigma_i^2}z$
    \For {$j=1,...,M$}
        \State $z \sim \mathcal{N}(0, I)$
        \State $x_i\leftarrow x_i+\epsilon_is_{\theta^\star}(x_i,\sigma_i) + \sqrt{2\epsilon_i}z$
    \EndFor
\EndFor\\
\Return $x_0$
\end{algorithmic}
\end{algorithm}

\begin{algorithm}
\caption{VP-SDE predictor-corrector sampling algorithm}\label{alg:vpsde-pc}
\begin{algorithmic}
\State $x_N \sim \mathcal{N}(0,I)$
\For {$i=N-1,...,0$}
    \State $x_i'\leftarrow (2-\sqrt{1-\beta_{i+1}})x_{i+1}+\beta_{i+1}s_{\theta^\star}(x_{i+1},i+1)$
    \State $z \sim \mathcal{N}(0, I)$
    \State $x_i\leftarrow x_i'+\sqrt{\beta_{i+1}}z$
    \For {$j=1,...,M$}
        \State $z \sim \mathcal{N}(0, I)$
        \State $x_i\leftarrow x_i+\epsilon_is_{\theta^\star}(x_i,i) + \sqrt{2\epsilon_i}z$
    \EndFor
\EndFor\\
\Return $x_0$
\end{algorithmic}
\end{algorithm}

\subsection{Relationship and limitations of the core studies} \label{subsection:limitations}
There are several key differences between noise conditional score networks\cite{song2019generative} and denoising diffusion probabilistic  models\cite{ho2020denoising}. NCSNs are trained with score-matching and sampled with Langevin dynamics, while DDPMs are trained with ELBO like VAEs and sampled with a trained decoder. Ho et al.\cite{ho2020denoising} mentions that the ELBO is essentially equivalent to a weighted combination of score-matching objectives in score-based generative modelling. They also demonstrated image quality that comparable or superior than GANs for the first time by parameterizing the decoder as an array of noise conditional score networks and using the U-Net architecture.

Score SDEs \cite{song2020score} proposed a framework that combines score-based generative models and diffusion probabilistic models. They generalized the number of noise scales to infinity and proved that both score-based generative models and diffusion models are different types of stochastic differential equations. 

Huang et al. \cite{huang2021variational} treats Brownian motion as a latent variable in order to track the log-likelihood estimation explicitly. This study fills the theoretical gap by showing that minimizing score matching loss is equivalent to maximizing ELBO with the reverse SDE that proposed by Song et al. \cite{song2020score}. Also Kingma and Gao \cite{kingma2023understanding} showed that all diffusion objectives are equal
to a weighted integral of ELBOs.

It is understood that score-based generative models and diffusion models are different perspectives of the same model family in line with these developments. In this study we will be refer this model family as generative diffusion models, or diffusion models in short, which is commonly used in the literature.

Diffusion models provide a high capacity of productivity and a tractable process but they have long sampling steps and their computation cost limits the applicability of these models. Moreover, some key factors in the training process affect the learning patterns and the performance of the models. The impact of these key factors should be explored to improve the model. The training objective of diffusion probabilistic models is variational lower bound. However, this objective is not tight in most cases, and it potentially leading to find a sub-optimal log-likelihood. Since log-likelihood is not minimized with the ELBO at the same time, some methods directly focus on the likelihood optimization problem. Therefore, likelihood maximization is a research topic for diffusion models. Also optimization of variational gap between ELBO and log-likelihood is a research topic.

Although diffusion models have had great success for data domains such as image and sound, they have problems in applications with other modalities. Many domains have special structures that must be considered to work with diffusion models effectively. For example, when models have score functions defined only in continuous space, or when data resides in low-dimensional manifolds difficulties may arise. To cope with these challenges, diffusion models must be generalized and adapted in various ways.

\section{Theoretical Research}
\begin{figure}[htbp]
    \centering
    \includegraphics[width=0.8\textwidth]{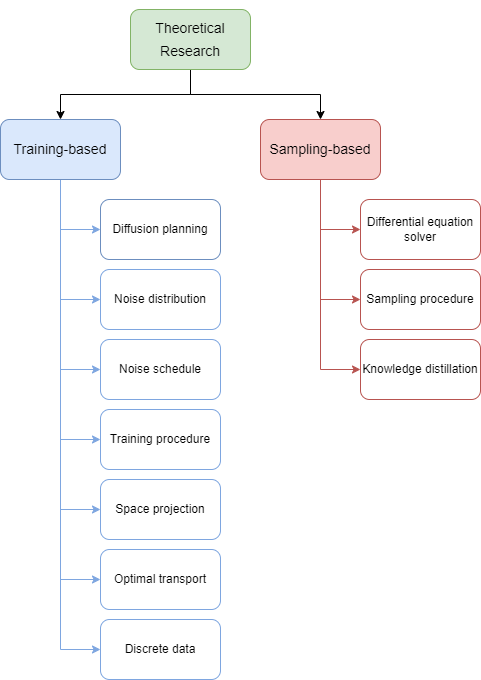}
    \caption{Categories of theoretical research}
    \label{fig:improvement}
\end{figure}

In this section, we survey improvements on the algorithm of diffusion models. We categorized the improvement studies under 2 main headings. The first one is the training-based approaches, the second one is the training-free sampling-based approaches. Under these categories we sub-sectioned the research according to the subjects they have focused. All sub-categories are shown in Figure \ref{fig:improvement}.

\subsection{Training-based approaches}
Improvements that change traditional training scheme have shown that key factors in the training process affect learning patterns and model performance. Training-based approaches are examined under 7 sub-headings which are diffusion planning, noise distribution \& schedule, training procedure, space projection, optimal transport and different data structures respectively.

\subsubsection{Diffusion planning}
In diffusion models, efficiently learning the forward diffusion process supports to optimize the sampling process indirectly. CCDF \cite{chung2022come} and Franzese et al. \cite{franzese2022much} proposed approaches that optimize the number of diffusion steps to minimize ELBO. 

Another perspective is based on truncation. The main idea is to early-stop the forward diffusion process and start the denoising process with a non-Gaussian distribution. Truncation operation have a truncation hyperparameter which balances sampling speed and sample quality. In the truncation, less diffused data is produced with the help of other generative models such as GAN and VAE. TDPM \cite{zheng2022truncated} samples from an implicit distribution found with GAN and conditional transport (CT)\cite{zheng2020act} to truncate both the diffusion and sampling process. ES-DDPM \cite{lyu2022accelerating} proposes the idea of early stopping for learning trajectories of the samples. ES-DDPM learns the necessary condition from the DiffuseVAE \cite{pandey2022diffusevae} instance to stop the sampling early.

Diffusion models contain an inverse correlation between density estimation and sampling performance. The small diffusion time contributes the density estimation significantly, while the large diffusion time is mainly improves the sample quality. Kim et al. \cite{kim2022soft} explores such an inverse correlation with sufficient empirical evidence in Soft Truncation. It is proposed as a training technique that smoothes the constant truncation hyperparameter to a random variable. The NCSN++ model is obtained by applying the method to noise conditional score networks and the DDPM++ model is obtained by applying it to denoising diffusion probabilistic models. Both models achieved state-of-the-art results.

\subsubsection{Noise distribution}
\begin{figure}[htbp]
    \centering
    \includegraphics[width=0.8\textwidth]{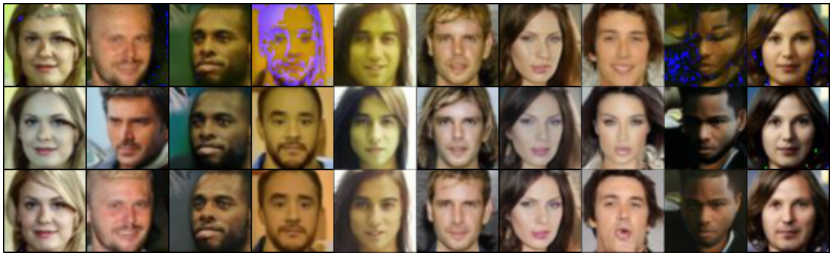}
    \caption{Results for different noise distributions. The first row is Gaussian noise, the second row is Mixture of Gaussian noise, and the third row is Gamma noise \cite{nachmani2021non}.}
    \label{fig:non-gaussian}
\end{figure}

The noise distribution of the diffusion process in the existing methods is Gaussian noise. However, if more degrees of freedom is given to the noise distribution, this can improve the performance. Nachmani et al. \cite{nachmani2021non, nachmani2021denoising} has applied non-Gaussian noise distributions and achieved better results with Gamma distribution for image and speech generation. This study shows using a mixture of Gaussian noises in the diffusion process improves the performance over a single distribution as shown in Figure \ref{fig:non-gaussian}. 

Xiao et al. \cite{xiao2021tackling} claims that the Gaussian assumption in the denoising process slowdowns the sampling speed, and this assumption only works for small step sizes. Their method allows for larger step size by modeling each step in the denoising process with a conditional GAN. It is the first DDPM model to produce samples in 4 steps with this method. Also, Non-uniform diffusion models \cite{batzolis2022nonuniform} and Non-isotropic gaussian noise models \cite{voleti2022scorebased} use different formulations to change standard uniform noising process.

Soft Diffusion \cite{daras2022soft} proposed Soft Score Matching to realize the noising process. This study has developed a new way to select noise levels and a new sampling method. Blurring Diffusion \cite{hoogeboom2022blurring} bridges the gap between standard Gaussian denoising diffusion and inverse heat dissipation \cite{rissanen2023generative}. They defined heat dissipation or blurring as a Gaussian diffusion process with non-isotropic noise. Also Chen et al. \cite{chen2023geometric} discover that the data distribution and the noise distribution are smoothly connected with two trajectories. The first one is an explicit, quasi-linear sampling trajectory, and the second one is an implicit denoising trajectory.

\subsubsection{Noise schedule}
Learning the noise scale can provide an explainable noising and a regular sampling process. There are some methods that consider the noise scale as a learnable parameter in reverse process as well as forward process. In existing methods, transition kernels in the reverse Markov chain are assumed to have constant covariance parameters, and these covariance values are hand-made without considering any conditions. In order to obtain more efficient sampling steps, the covariance optimization can be performed dynamically and the noise scale in the reverse process can be learned in this way. 

Improved DDPM \cite{nichol2021improved} proposes to parameterize the reverse covariances as a linear interpolation and use a hybrid objective function. They parameterize the reverse covariance as in Equation \ref{eqn:reverse_variance} and obtained higher log-likelihoods and a better sampling speed without compromising sample quality.

\begin{equation}
\Sigma_\theta(x_t,t)=e^{\theta\cdot log\beta_t + (1-\theta)\cdot log\overline{\beta}_t}
    \label{eqn:reverse_variance}
\end{equation}

Improved DDPM has demonstrated that the cosine noise schedule given in Equation \ref{eqn:iddpm} can improve log probabilities. Here $s$ is a given hyperparameter to control the noise scale at time $t=0$.

\begin{equation}
\overline{a}_t=\frac{h(t)}{h(0)}, \quad   h(t)=cos\Big(\frac{t/t + s}{1 + s}\cdot \frac{\pi}{2}\Big)
    \label{eqn:iddpm}
\end{equation}

FastDPM \cite{kong2021fast}, examines fast sampling methods for different domains, different datasets, and different amounts of information for conditional generation. FastDPM connects the noise design with the ELBO optimization, deriving from the variance constant for DDPM or the time step in DDIM \cite{song2020denoising}. In addition, this study demonstrated that the performance of the methods depends on the domain (for example, image or sound), the balance between sampling speed and sample quality, and the amount of conditional information. There are also recommendations for those undecided what method to use.

DiffFlow \cite{zhang2021diffusion} runs the noise addition process by minimizing the KL-Divergence between the forward and backward process with a flow function at each SDE-based diffusion step. This method requires a longer time per step due to backpropagation of the flow functions, but is still 20 times faster than DDPM as it needs fewer steps. DiffFlow learns more general distributions with fewer discretization steps compared to DDPM.

Variational Diffusion Models (VDM) \cite{kingma2021variational} has improved the log-probability of continuous-time diffusion models by training the noise schedule and other diffusion model parameters together. The noise process is parameterized by using a monotonous neural network $\gamma_\eta(t)$. Forward noise process is designed as $\sigma_t^2=sigmoid(\gamma_\eta(t)), q(x_t|x_0)=\mathcal{N}(\overline{a}_tx_0, \sigma_t^2I)$ and $\overline{a}_t=\sqrt{(1-\sigma_t)^2}$. Also, the authors note that for any $x$ data point, the VLB can be simplified to a form that depends only on the signal-to-noise ratio $R(t):=\frac{\overline{a}_t^2}{\sigma_t^2}$. In the last case $L_{VLB}$ can be represented as in Equation \ref{eqn:vdm_1}. Here the first and second terms can be directly optimized like VAEs. The third term is given in Equation \ref{eqn:vdm_2}.

\begin{equation}
L_{VLB}=-E_{x_0}\text{KL}(q(x_T|x_0) || p(x_T)) + E_{x_0,x_1}logp(x_0|x_1)-L_D
    \label{eqn:vdm_1}
\end{equation}

\begin{equation}
L_D=\frac{1}{2}E_{x_0, \epsilon \sim \mathcal{N}(0,I)}\int_{R_{min}}^{R_{max}}||x_0-\overline{x}_\theta(x_\nu,\nu)||_2^2d\nu
    \label{eqn:vdm_2}
\end{equation}

$R_{max}=R(1)$ ve $R_{min}=R(T)$ in Equation \ref{eqn:vdm_2} specifies the noisy data point obtained from forward diffusion until $x_\nu=\overline{a}_\nu x_0 + \sigma_\nu\epsilon$ $x_0$'ın $t=R^{-1}(\nu)$. $\overline{x}_\theta$ indicates the denoised data point which predicted by the diffusion model. Noise processes do not affect the VLB as $R_{min}$ and $R_{max}$ share the same values. This parameterization only affects the variance of the Monte Carlo estimators.

Bilateral denoising diffusion models (BDDMs) \cite{lam2022bddm} learns a score network and a scheduling network together and speed up the sampling process by optimizing the inference noise schedules. Dynamic Dual-Output Diffusion Models \cite{benny2022dynamic} have two opposite equations for the denoising process, the first one is estimating the applied noise, and the second one is estimating the image. It dynamically alternate between them to obtain efficiency in terms of quality and speed. Also San Roman et al. \cite{sanroman2021noise} controlled the denoising process with a seperate neural network and obtained a noise schedule based on the conditioning data.

Lin et al. \cite{lin2023common} rescale the noise schedule and enforce the last step to have zero signal-to-noise ratio (SNR) and solve the problem of always generating images with medium brightness in Stable Diffusion \cite{rombach2022high}. 

Optimal linear subspace search (OLSS) \cite{duan2023optimal} proposed a faster scheduler by searching for the optimal linear subspace. Chen et al. \cite{chen2023importance} found that noise schedule should depend on the image size. Increasing the size, a noisier schedule should be applied. Also Choi et al. \cite{choi2022perception} obtained better results by redesigning the weighting scheme and giving priority to some noise levels.

Another perspective on the denoising process is proposed in the Cold Diffusion \cite{bansal2022cold} study. Cold diffusion claims that the generative behavior of diffusion models is not strongly dependent on noise selection but a whole family of generative models can be designed by changing this selection. They used a completely deterministic denoising process and obtained other models by generalizing the update rules.

\subsubsection{Training procedure}
Some approaches suggest different training strategies to reduce the computation cost and achieve better generative quality.

To reduce the cost of directly maximizing $p_\theta^{\text{ODE}}$ Song et al. \cite{song2021maximum}, suggest to maximize the variational lower bound of $p_\theta^{\text{SDE}}$ instead of $p_\theta^{\text{ODE}}$ and introduced a family of diffusion models called ScoreFlows. ScoreFlows run the diffusion process using the normalizing flow to transform the data distribution into a dequantization field and generate dequantized samples. Operating the process in the dequantization field with variational dequantization resolves the incompatibility between continuous and discrete data distributions. 

Lu et al. \cite{lu2022maximum} further enhances ScoreFlows by proposing to minimize not only the score matching loss, but also higher order generalizations. They proved that $logp_\theta^{\text{ODE}}$ can be constrained by first, second, and third order score matching losses. On top of that, they were able to improve $p_\theta^{\text{ODE}}$ with their proposed efficient training algorithms which minimize high-order score matching losses. They explore that there is a gap between ODE likelihood and score matching objectives. They also prove that first-order score matching is not enough to maximize the likelihood of ODE. They propose a new high-order denoising score matching method to enable maximum likelihood training of score-based diffusion ODEs. They tracked the first, second, and third order score matching errors and limit the negative likelihood of ODE. Then they obtained high likelihood and sample quality with this method.

\begin{figure}[htbp]
    \centering
    \includegraphics[width=0.8\textwidth]{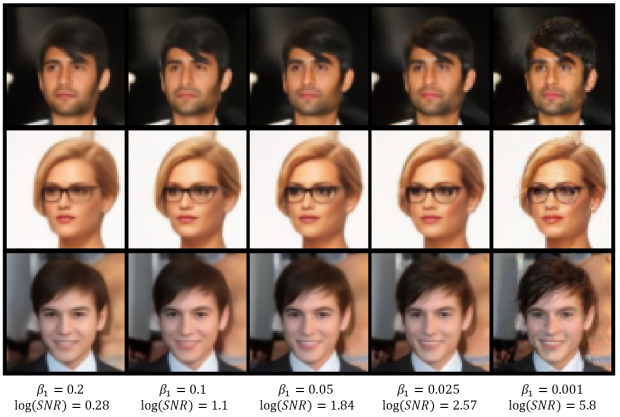}
    \caption{Examples from DAED with different switching points($\beta$) \cite{deja2022analyzing}.}
    \label{fig:switching-points}
\end{figure}

Denoising Auto-Encoder with Diffusion(DAED) \cite{deja2022analyzing} have a switching point and divide the denoising process into two parts: a denoising part and a generating part and used different procedures in these parts. Figure \ref{fig:switching-points} shows examples from DAED with the
same noise and different switching points where the higher $\beta$ means long term noise. Also, Multi-architecturE Multi-Expert (MEME) \cite{lee2023multiarchitecture} have different denoiser experts in the noise intervals to consider the optimal operations for each time-step.

Cho et al. \cite{cho2023enhanced} trained a diffusion model with two latent codes, a spatial content code and a style code. They also proposed two novel techniques to improve controllable sampling.

Yi et al. \cite{yi2023generalization}  emphasis on the generalization ability of the diffusion models and proposed a novel objective which has no generalization problem. Also Fast Diffusion Model (FDM) \cite{wu2023fast} speeds up both training and sampling from a stochastic optimization perspective.

Phoenix \cite{jothiraj2023phoenix} used federated learning to train an unconditional diffusion model
to improve the data diversity even the data with statistical heterogeneity or Non-IID (Non-Independent and Identically Distributed). Ning et al. \cite{ning2023input} regularized the training by perturbing the ground truth samples to alleviate exposure bias problem. Also, WaveDiff \cite{phung2022wavelet} takes each input as four frequency sub-bands (LL, LH, HL, HH) and reduced both training and inference times.

ProDiff \cite{huang2022prodiff} has been used knowledge distillation \cite{hinton2015distilling} for text-to-sound generation. Here, the students distill knowledge directly from scratch, by minimizing KL-divergence between two categorical distributions. 

For the purpose of conditional generation Classifier-free guidance \cite{ho2022classifier} jointly train a conditional and an unconditional diffusion model, and combine their predicted scores to trade-off the diversity for the quality of the samples. Pyramidal diffusion model \cite{ryu2022pyramidal} trained a score function with a positional embedding and generated high-resolution images from low-resolution inputs. 

CARD \cite{han2022card} combines a denoising diffusion model with a pre-trained conditional mean estimator to predict data distribution under given conditions. Direct Optimization of Diffusion Latents (DOODL) \cite{wallace2023endtoend} suggest to optimize the gradients of a pre-trained classifier on the true generated pixels for memory-efficient backpropagation. Also, Model Predictive Control (MPC) \cite{shen2022conditional} guide the diffusion process by backpropagating the explicit guidance in additional time-steps. 

Blattmann et al. \cite{blattmann2022semiparametric, blattmann2022retrieval} retrieve a set of nearest neighbors from an external database and use these samples to condition the model. There are also some studies \cite{sheynin2022knndiffusion, rombach2022textguided, chen2022reimagen} which used retrieval-augmented methods to improve the generative performance in specific tasks. Also, Self-guided diffusion model \cite{hu2023selfguided} give attention to the cost of obtaining image-annotation pairs for the guidance and eliminate the need for annotation by using self-supervision signals. 

Also Cascaded Diffusion Model \cite{ho2022cascaded} is a chain of diffusion models that generates images in increasing resolution and they reported that conditioning augmentation provides to achieve state-of-the-art FID scores in conditional generation tasks.

\subsubsection{Space projection}
Mapping a high-dimensional input to a high-dimensional output at every step in the reverse process increases computation cost. Projecting data to a low-dimensional space is one of the first solutions that come to mind to speed up the model. Subspace Diffusion \cite{jing2022subspace} and Dimensionality-varying diffusion process (DVDP) \cite{zhang2023dimensionality} projected the input into a smaller subspace in the forward diffusion process. 

In the Critically-damped Langevin Diffusion (CLD)\cite{dockhorn2021score}, the process is in an expanded space where the variables are considered as "velocities" which depend on the data. To obtain the reverse-SDE with this method, it is sufficient to learn the score function of the conditional distribution of the "velocities" rather than learning the data scores directly. This method improves the sampling speed and quality.

Diffusion models are based on transition kernels that must be in equivalent spaces and this prevents dimension reduction. For this reason, diffusion models can be trained in latent space using an autoencoder. However, it is necessary to use a loss function that allows to train the autoencoder and the diffusion model jointly. 

\begin{figure}[htbp]
    \centering
    \includegraphics[width=0.8\textwidth]{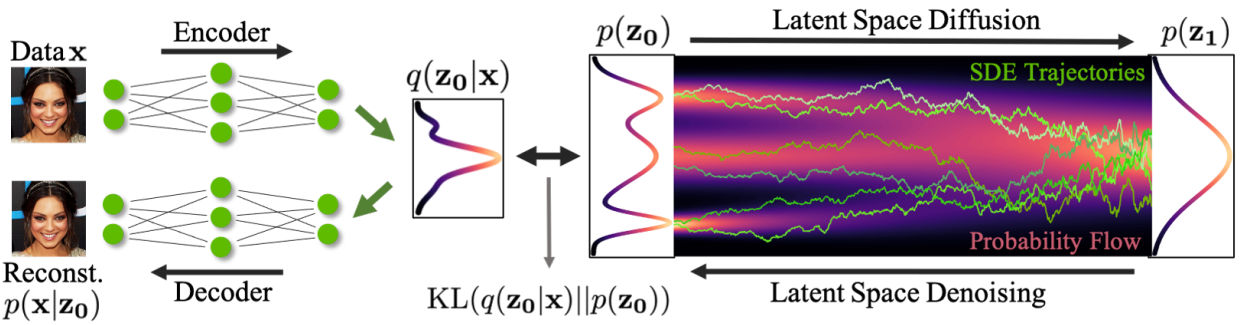}
    \caption{Mechanism of the latent score-based generative model (LSGM) \cite{vahdat2021score}.}
    \label{fig:lsgm}
\end{figure}

Latent Score-Based Generative Model (LSGM) \cite{vahdat2021score} in Figure \ref{fig:lsgm}, maps the data to the latent space with an encoder and the diffusion process is applied in the latent space. Then, then samples in the latent space are mapped to data space using a decoder. LSGM also demonstrated that ELBO is a specific score-matching objective for latent space diffusion. Thus the intractable cross-entropy term in ELBO can be calculated by converting it to the score matching objective. They optimized a new lower bound for log-likelihood by combining the evidence lower bound (ELBO) of the VAE and the score matching objective of the diffusion model in the LSGM. LSGM generates samples faster than traditional diffusion models because it projects the training samples into the latent space with a lower dimension. In addition, LSGM can project discrete data into latent space to convert and use them in continuous space.

\begin{figure}[htbp]
    \centering
    \includegraphics[width=0.8\textwidth]{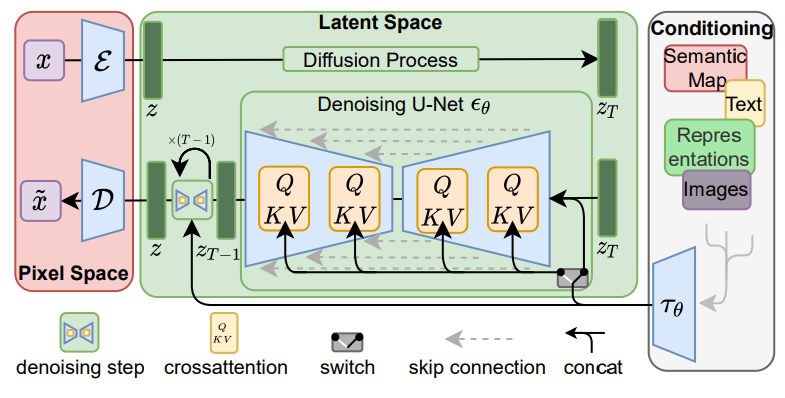}
    \caption{Mechanism of the latent diffusion model (LDM) \cite{rombach2022high}.}
    \label{fig:ldm}
\end{figure}

On the other hand, Latent Diffusion Model (LDM) \cite{rombach2022high} in Figure \ref{fig:ldm} trained the autoencoder and obtained a low-dimensional latent space firstly. Then, the diffusion model generated the latent codes in this latent space. LDM is able to build the UNet from 2D convolutional layers so that the reweighted objective focus on the most relevant bits. Augmenting the UNet with the cross-attention mechanism becomes with the effectiveness in flexible conditional generation. Stable Diffusion is a latent diffusion model, and one of the most frequently used text-to-image generator.

DiffuseVAE \cite{pandey2022diffusevae} uses VAEs for conditional generation and succeeded on controllable synthesis in a low-dimensional latent space. They use a generator-refiner framework. In this framework a conditioning signal (y) is first modeled using a standard VAE. Then they model the data using a DDPM conditioned on y and the low-dimensional VAE latent code representation of y. Also Pandey et al. \cite{pandey2023generative} constructed the forward process in a space augmented with auxiliary variables and proposed Phase Space Langevin Diffusion (PSLD).

Xu et al. project the data points as electrical charges to an augmented space in Poisson flow generative models (PFGM) \cite{xu2022poisson} and extend their study to the large scale datasets in PFGM++ \cite{xu2023pfgm++} by unifying with diffusion models.

Some space projection methods apply data transformation to normalizing flows.  Parametrized Diffusion Model (PDM) \cite{kim2021maximum} has obtained a faster computation by finding latent variables with the flow function. In addition, they define the variational gap expression which corresponding to the gap between ELBO and log-likehood optimizations. They also propose a solution to eliminate the gap by collective learning. Implicit Non-linear Diffusion Model (INDM) \cite{kim2022maximum} applies the flow model to express the variational gap and minimizes the gap by training the bidirectional flow model and the linear diffusion model on the latent space jointly. While INDM \cite{kim2022maximum} and PDM\cite{kim2021maximum} build models in a smaller space, this results in less evaluation steps and faster sampling.

CycleDiffusion \cite{wu2022unifying} have a latent space which is emerged from two diffusion models trained independently on related domains. Also, Multi-modal Latent Diffusion \cite{bounoua2023multimodal}, uni-modal autoencoders are independently trained,  and their latent variables are combined in a common latent space to use in a diffusion model.

Boomerang \cite{luzi2022boomerang} moves an input image to the latent manifold space by perturbing, then it comes back to the image space through local sampling on image manifolds.

Jump diffusion process \cite{campbell2023transdimensional} jumps between dimensional spaces by destroying the dimension in the forward process and creating again in the reverse
process with a novel ELBO learning.

\subsubsection{Optimal transport}
De Bortoli et al. \cite{de2021simulating} simulated diffusion bridges whose start and end states are conditional diffusion processes. It uses a variational score mapping formulation to learn the reverse transformation with the functions. In a recent study Khrulkov et al. \cite{khrulkov2023understanding} showed that the DDPM encoder map coincides with the Monge optimal transport map for multivariate normal distributions.

Su et al. \cite{su2022dual} defined Dual Diffusion Implicit Bridges (DDIB) as a concatenation of source to latent, and latent to target Schrodinger Bridges, which is a form of entropy-regularized optimal transport.

Another method was proposed by Lee et al. \cite{lee2023minimizing} which claimed that high curvature of the learned generative trajectories reason with slow sampling speed. They proposed to train the forward process to minimize the curvature of generative trajectories and decreased the sampling cost with competitive performance.

Zheng et al. \cite{zheng2023fast} proposed diffusion model sampling with neural operator (DSNO) method which maps the Gaussian noise distribution, to the trajectory of the reverse diffusion process in the continuous-time solution. They used neural operators to solve the probability flow ODE and achieved state of-the-art FID scores.

$\alpha$-blending \cite{heitz2023iterative} showed that linearly interpolating (blending) and deblending samples iteratively produces a deterministic mapping between the two densities and this mapping can be used to train a neural network which is a minimalist deterministic diffusion model that deblend samples.

\begin{figure}[htbp]
    \centering
    \includegraphics[width=0.8\textwidth]{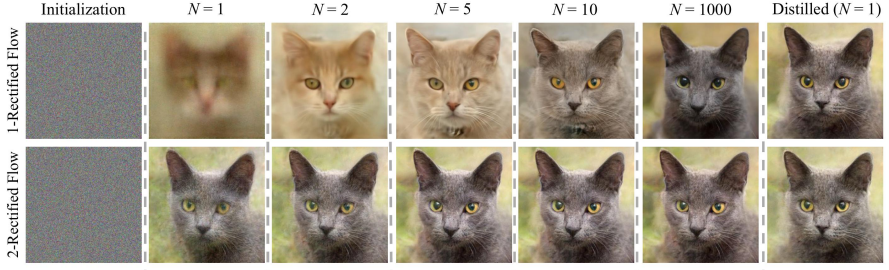}
    \caption{Rectified flow \cite{liu2022flow} generates good samples with a very small number of steps (N $\geq$ 2) in image generation task.}
    \label{fig:recflow}
\end{figure}

The Rectified flow \cite{liu2022flow} is an ODE model that transport the noise distribution to the data distribution by following straight line shortest paths which found with least squares optimization as shown in Figure \ref{fig:recflow}. They obtain high quality results only with a single Euler discretization step. 

Stochastic Interpolant \cite{albergo2022building} is a normalizing flow model between any base and target pair of distributions. The velocity of this model can be used to construct a diffusion model that estimate the data score and sample from it. Flow Matching \cite{lipman2022flow} is also suitable for transforming noise to data samples for Gaussian probability paths.

\subsubsection{Discrete data}
Since Gaussian noise addition is not compatible for discrete data, most diffusion models can only be applied for continuous data. Moreover, score functions are defined only for continuous data. However, it is necessary to work with discrete data such as sentences, atoms/molecules and vectorized data in machine learning problems. For this purpose, there are some methods that produce high-dimensional discrete data with diffusion models. 

The sampling process of autoregressive models is time-consuming, especially for high-dimensional data, as it must be sequential. Autoregressive Diffusion Models (ARDM) \cite{hoogeboom2021autoregressive} is trained with an objective that uses probabilistic diffusion models instead of using causal masking in representations. ARDM can generate data in parallel during the test stage, which allows to apply it to the conditional generation tasks.

\begin{figure}[htbp]
    \centering
    \begin{subfigure}[b]{0.8\textwidth}
    \centering
    \includegraphics[width=\textwidth]{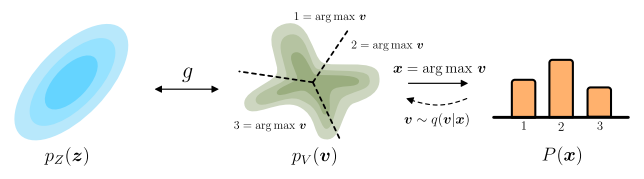}
    \caption{Argmax flows}
    \label{fig:argmax}
    \end{subfigure}
    
    \begin{subfigure}[b]{0.8\textwidth}
    \centering
    \includegraphics[width=\textwidth]{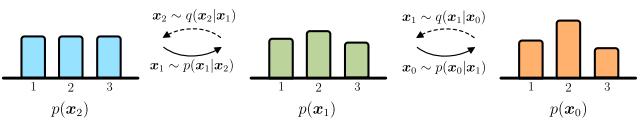}
    \caption{Multinomial diffusion}
    \label{fig:multinomial}
    \end{subfigure}
    \caption{Two extensions of flows
and diffusion by Hoogeboom et al. \cite{hoogeboom2021argmax}}
\end{figure}

For learning categorical distributions Hoogeboom et. al. \cite{hoogeboom2021argmax} introduces two extensions to the flow-based models and diffusion models. The extension to flows is combination of a continuous distribution (like a normalizing flow) and an argmax function which shown in Figure \ref{fig:argmax}. This model learns to optimize the probabilistic inverse of Argmax which moves categorical data into a continuous space. On the other hand, multinomial diffusion in Figure \ref{fig:argmax} adds categorical noise gradually during the diffusion process in extension to diffusion models. The proposed method outperforms existing approaches in text modeling and image segmentation maps in terms of log-likelihood optimization.

Discrete denoising diffusion probability models (D3PM) \cite{austin2021structured}, generalized the multinomial diffusion model. This study establishes a connection between diffusion models and autoregressive models. D3PM performs the forward noise processing with absorbing state kernels or discretized Gaussian kernels to adapt diffusion models to discrete data. It shows that the selection of the transition matrix is an important design decision that improves the results in image and text domains. It also introduces a new loss function that combines the variational lower bound with cross-entropy loss. 

Gu et al. \cite{gu2022vector} first used diffusion models for vector quantized (VQ) data. In this study, they solved the unidirectional bias and the accumulation prediction error problems in VQ-VAEs\cite{van2017neural}. VQ-Diffusion \cite{gu2022vector}, uses a random walk or masking operation in the discrete data space instead of Gaussian noise.

Analog Bits \cite{chen2023analog} represent the discrete data as binary bits, and model these bits with a continuous diffusion model. They reported that their two novel techniques (Self-Conditioning and Asymmetric Time Intervals) improve the sample quality significantly. Blackout Diffusion \cite{santos2023blackout} generalize Gaussian forward processes to discrete-state processes and generate samples from an empty image instead of noise. Also Campbell et al. \cite{campbell2022continuous} outperform their discrete counterparts by using continuous-time markov chains.

\subsection{Sampling-based approaches}
In this section we will examine the approaches which directly improve the sampling algorithm with pre-trained models. Sampling-based approaches are categorized under 3 sub-headings. These are differential equation solver, sampling procedure, and knowledge distillation, respectively.

\subsubsection{Differential equation solver}
High-order differential equation solvers have higher degree of convergence and lower estimation errors but requires more computational costs and has instability issues. There are two basic differential equation formulations in general: The SDE formulation walks in a random direction in the noise field, on the other hand, ODE formulation has high speed since it is deterministic. 

Song et al. \cite{song2021maximum} proves that the objective used for training Score-SDEs maximizes the expected value of the generated distribution with a custom weighting function (likelihood weighting). If the distribution generated by SDE is denoted by $p_\theta^{\text{SDE}}$, the objective function can be given as in Equation \ref{eqn:likelihood_SDE}. Here $\mathcal{L}(\theta;g(.)^2)$, is the objective function given in Equation \ref{eqn:sde_fisher}, and $\lambda(t)=g(t)^2$. 

\begin{equation}
D_{KL}(q_0 || p_\theta^{\text{SDE}}) \leq \mathcal{L}(\theta;g(.)^2) + D_{KL}(q_t||\pi)
    \label{eqn:likelihood_SDE}
\end{equation}

$D_{KL}(q_0||p_\theta^{\text{SDE}})=-E_{q_0} log(p_\theta^{\text{SDE}})+\text{const}$, and $D_{KL}(q_t||\pi)$ is a constant. Training $\mathcal{L}(\theta;g(.)^2)$ minimizes the negative log-likelihood on data $-E_{q_0}log(p_\theta^{\text{SDE}})$. Equation \ref{eqn:likelihood_bound} shows the bound applied for $p_\theta^{\text{SDE}}(x)$\cite{huang2021variational,song2021maximum}.

\begin{equation}
-log(p_\theta^{\text{SDE}}(x) \leq \mathcal{L}'(x)
    \label{eqn:likelihood_bound}
\end{equation}

Here $\mathcal{L}'(x)$ is calculated as in the Equation \ref{eqn:bound_loss}. The first part of this loss reminds the implicit score-matching \cite{hyvarinen2005estimation}. Equation \ref{eqn:bound_loss} can be estimated efficiently by Monte Carlo methods.

\begin{equation}
\mathcal{L}'(x) = \int_0^TE\Big[\frac{1}{2}||g(t)s_\theta(x_t,t)||^2 + \nabla\cdot(g(t)^2s_\theta(x_t,t)-f(x_t),t) \Big|x_0=x \Big]dt - E_{x_T}[logp_\theta^{\text{SDE}}(x_T)| x_0=x] 
    \label{eqn:bound_loss}
\end{equation}

Since probability flow ODE is a special case of neural ODEs or continuous normalizing flows, the distribution generated by the ODE $p_\theta^{\text{ODE}}$ is calculated over these approximations in Equation \ref{eqn:likelihood_ODE}.

\begin{equation}
logp_\theta^{\text{ODE}}(x_0)=logp_T(x_T)+\int_0^T\nabla\cdot\overline{f}_\theta(x_t,t)dt
    \label{eqn:likelihood_ODE}
\end{equation}

The single integral here can be calculated with numerical ODE solvers and the Skilling-Hutchinson trace estimator \cite{hutchinson1989stochastic, skilling1989eigenvalues}. However, this formula cannot be directly optimized to maximize $p_\theta^{\text{ODE}}$ because it requires calling expensive ODE solvers for each data point $x_0$. 

Ito-Taylor Sampling Schedule \cite{tachibana2021taylor} improves accuracy by using a high-order solver. In this method, the score function is parameterized to avoid the calculation of high-order derivatives. Dockhorn et al. \cite{dockhorn2022genie} selected the high-order score functions of the first order score network and trained another network to speed up sampling by calculating required high-order terms only.

Speed-oriented methods improve the whole process using both linear solvers and high-order solvers. Jolicoeur-Martineau et al. \cite{jolicoeur2021gotta} has developed an SDE solver with adaptive step sizes for faster generation. A linear solver (Euler-Maruyama Method) is combined with a high-order solver (Improved Euler Method) to speed up the sampling process. The high and low order solvers generate new $x_{high}$ and $x_{low}$ samples from the previous $x_{prev}$ sample at each time step. Then the step size is adjusted by comparing the difference between these two samples. If $x_{high}$ and $x_{low}$ are similar, the algorithm returns $x_{high}$ and increases the step size. The difference between $x_{high}$ and $x_{low}$ is given in Equation \ref{eqn:jol-mar}. Here $\delta(x,x_{prev}):=max(\epsilon_{abs}-\epsilon_{rel}max(|x|,|x_{prev}|))$, $\epsilon_{abs}$ and $\epsilon_{rel}$ are absolute and relative tolerances.

\begin{equation}
E_q=\Big|\Big|\frac{x_{high}-x_{low}}{\delta(x,x_{prev})}\Big|\Big|^2
    \label{eqn:jol-mar}
\end{equation}

The Diffusion Exponential Integrator Sampler (DEIS) \cite{zhang2022fast} take advantage of the semi-linear nature of probability flow ODE to develop more efficient customized ODE solvers. The linear part of the probability flow ODE is calculated analytically, and the nonlinear part is solved with exponential integrators. In these methods, DDIM\cite{song2020denoising} is the first-order estimator, then the high-quality samples are generated in only 10-20 iterations by using high-order integrators. Also, DPM-Solver \cite{lu2022dpm} uses different order solvers. DPM Solver-Fast has obtained better performance  with the solvers combined in alternative cross-orders. Also Pang et al. \cite{pang2023calibrating} proposed a novel method for calibrating a pre-trained DPM-solver\cite{lu2022dpm} model and obtained better results in discrete-time models.

PNDM \cite{liu2022pseudo} solves the differential equation during sampling with the help of manifold space. It provides a generalized version for the differential equation sampler. PNDM has a pseudo-numerical method to sample from a given manifold in $R^N$ assuming that different numerical solvers can share the same gradient. They use a three-step high-level solver (Runge-Kutta method), then a multi-step linear method for sampling.

EDM \cite{karras2022elucidating} proposes a "churn" step in the sampling process. It enhances the corrector of Predictor-Corrector sampler \cite{song2020score}. EDM uses the time steps of deterministic diffusion ODE with the second-order Heun solver\cite{ascher1998computer}. They claim that the 2nd order Heun solver provides a perfect balance between sample quality and sampling speed. Heun's method produces competitive samples with fewer sampling steps. 

ODE-based samplers are fast while SDE-based samplers generate high quality samples. Restart \cite{xu2023restart} alternates between adding substantial noise in additional forward steps and strictly following a backward ODE. Restart outperforms previous samplers both in terms of calculation time and quality. 

Cao et al. \cite{cao2023exploring} explored to choice SDE or ODE based diffusion models and found that when they perturb the distribution at the end, the ODE model outperforms the SDE model but when they perturb earlier, the SDE model outperforms the ODE model.

Verma et al. \cite{verma2023variational} provides an alternative parameterization for Variational gaussian
processes by using continuous exponential SDEs and obtained a fast algorithm with fixed-point iterations for convex optimization. Score-integrand Solver (SciRE-Solver) \cite{li2023sciresolver} accelerates sampling with the help of recursive difference(RD). Also Meng et al. \cite{meng2021estimating} generalized denoising score matching to higher order derivatives.

\subsubsection{Sampling procedure}
There are many approaches that change the traditional sampling procedure and obtained better results without retraining the model. Implicit samplers are a class of step-skip samplers that do not require retraining the diffusion model. The number of steps in the diffusion process is usually equal to the number of steps in the sampling process. However, since the diffusion and sampling processes are separate, it is not a necessity. Song et al. \cite{song2020denoising} used a deterministic forward process and step-skip sampling in DDIMs. They designed non-Markovian diffusion processes so that DDIM is able to produce high-quality samples with deterministic processes much faster.

Obtaining $x_{t-1}$ sample from the $x_t$ sample is given in the equation \ref{eqn:ddim}. Here $\epsilon_t\sim\mathcal{N}(0,I)$, is standard Gaussian noise independent of $x_t$ and defined as $a_0:=1$. Different choices of $\sigma$ values result in different generative processes, all of these generative processes use the same $\epsilon_\theta$ pattern, therefore retraining is not required. If $\sigma_t=\sqrt{\frac{1-a_{t-1}}{1-a_t}}\sqrt{1-\frac{a_t}{a_{t-1}}}$ the forward process becomes Markovian and the model becomes DDPM. In case where $\sigma_t=0$ for all $t$, since $x_{t-1}$ and $x_0$ are known, the process becomes deterministic except $t=1$. The coefficient in front of the random noise $\epsilon_t$ in the generative process becomes zero, and samples are generated from the latent variables by a fixed procedure in an implicit probabilistic model.

\begin{equation}
x_{t-1}=\sqrt{a_{t-1}}\Big(\frac{x_t-\sqrt{1-a_t}\epsilon_\theta^{(t)}(x_t)}{\sqrt{a_t}}\Big)+\sqrt{1-a_{t-1}-\sigma_t^2}\cdot\epsilon_\theta^{(2)}(x_t)+\sigma_t\epsilon_t
    \label{eqn:ddim}
\end{equation}

When the iteration of DDIM is rewritten as in Equation \ref{eqn:ddim_rw}, it is similar to Euler integration for solving ordinary differential equations (ODEs). Here,
$\frac{\sqrt{1-a}}{\sqrt{a}}$, $\sigma_t$ and, $\frac{x}{\sqrt{a}}$ is parametrized with $\overline{x}$ then the corresponding ODE is derived. In the continuous case $\sigma$ and $\overline{x}$ are functions of $t$ and $\sigma(0)=0$.

\begin{equation}
\frac{x_{t-\Delta t}}{\sqrt{a_{t-\Delta t}}}=\frac{x_t}{\sqrt{a_t}}+\Big(\sqrt{\frac{1-a_{t-\Delta t}}{a_{t-\Delta t}}}-\sqrt{\frac{1-a_t}{a_t}}\Big)\epsilon_\theta^{(t)}(x_t)
    \label{eqn:ddim_rw}
\end{equation}

Equation \ref{eqn:ddim_rw} is the Euler method of the ODE given in Equation \ref{eqn:ddim_ODE}. Initial conditions $x(T)=\mathcal{N}(0,\sigma(T))$ correspond to $a\approx0$ for very large $\sigma_T$. This shows that the generative process can also be reversed with sufficient discretization steps. Encoding $x_0$ to $x_T$ simulates the reverse-ODE in Equation \ref{eqn:ddim_ODE}.

\begin{equation}
d\overline{x}(t)=\epsilon_\theta^{(t)}\Big(\frac{\overline{x}(t)}{\sqrt{\sigma^2+1}}\Big)d\sigma_t
    \label{eqn:ddim_ODE}
\end{equation}

The authors also refer to the concurrent study \cite{song2020score}, implying that the DDIM sampling process corresponds to VE-SDE, a special discretization of the probability flow ODE.

The generalized DDIM (gDDIM) \cite{zhang2022gddim} has generalized all of the implicit sampling diffusion models to a DDIM family with various kernel types in the SDE framework. They obtain a high sampling speed with the critically-damped Langevin diffusion (CLD) \cite{dockhorn2021score} model and a multi-step exponential sampler.

Analytic-DDPM \cite{bao2022analytic} finds the reverse covariance based on the reverse mean at each step. Equation \ref{eqn:analytic_DPM} shows obtaining the optimal reverse covariance from a pre-trained score model in the analytical form. When a pre-trained score model is given, estimating first and second order moments to obtain optimal reverse covariances and using them in the objective function results in tighter variational lower bound and higher probability values. A later study by the same authors \cite{bao2022estimating} proposes to estimate covariance by training another network on pre-trained DDPM models. In both of these studies, implementations were performed for both DDPM and DDIM models.

\begin{equation}
\Sigma_\theta(x_t,t)=\sigma_t^2 + \Big(\sqrt{\frac{\overline{\beta}_t}{a_t}}-\sqrt{\overline{\beta}_{t-1}-\sigma_t^2}\Big)^2 \cdot \Big(1-\overline{\beta}_tE_{q_t(x_t)}\frac{||\nabla_{x_t}logq_t(x_t)||^2}{d}\Big)
    \label{eqn:analytic_DPM}
\end{equation}

Efficient Sampling \cite{watson2021learning}, reorganise the objective for dynamic programming when a pre-trained diffusion model is given. In this method choosing the appropriate time schedules in the sampling process is an optimization problem. They decompose ELBO into individual KL terms and found the sampling trajectory that maximizes ELBO. The authors confirm that the ELBO optimization does not match the FID scores and recommend to explore another way to optimize the sampling trajectory. In a later study from same authors, Generalized Gaussian Diffusion Models (GGDM) \cite{watson2021learningiclr}, directly optimize Kernel Inception Distance (KID) \cite{binkowski2018demystifying} by using Differential Diffusion Sampler Search (DDSS). This study generalizes diffusion models with non-Markovian samplers and a wide range of marginal variance. This method uses reparameterization and gradient rematerialization tricks for backpropagation in the sampling process and brings memory cost instead of computation time.

Exposure bias problem is the mismatch of the inputs in training and inference processes. Training process always uses real samples, while the inference process uses the previously generated noisy sample. Real and noisy samples cause discrepancy when they given to the noise predictor network, which leads to error accumulation and sampling drift. Ning et al. \cite{ning2023elucidating} showed that practical sampling distribution has a larger variance than the ground truth distribution at every single step. They use a metric to evaluate the variance difference
between two distributions and proposed  Epsilon Scaling (ES). Also Time-Shift Sampler \cite{li2023alleviating} alleviates the exposure bias problem by adjusting the next step according to the variance of the current samples. Xu et al. \cite{xu2023stable} calculated weighted conditional scores for more stable training targets and this results in trading bias for reduced variance.

ParaDIGM \cite{shih2023parallel} parallelize the sampling process by estimating the solution of the sampling steps and tuning iteratively until convergence. Parameter-Efficient Tuning \cite{xiang2023closer} suggests that the input parameters holds the critical factor for the performance of downstream tasks. Best results have obtained by putting the input block after a cross-attention block. Aiello et al. \cite{aiello2023fast} 
minimized Maximum Mean Discrepancy
(MMD) to finetune the learned distribution with a given budget
of timesteps. Also Diff-Pruning \cite{fang2023structural} is an efficient compression method for learning lightweight diffusion models from pre-trained models, without re-training.

In the context of conditional generation, Dhariwal \& Nichol \cite{dhariwal2021diffusion} proposed classifier guidance as a post-training method to boost the sample quality. They combine the score of the pre-trained diffusion model with the gradient of another image classifier to condition an unconditional model. Graikos et al. \cite{graikos2023conditional} adapt a pre-trained unconditional model to the conditions by utilizing the learned representations of the denoising network. Also, Kawar et al. \cite{kawar2023enhancing} utilized the gradients of adversarially robust classifiers to guide the diffusion based image synthesis.

Discriminator Guidance \cite{kim2023refining} improved pre-trained models by utilizing from an explicit discriminator network which predicts whether the produced sample is realistic or not. FSDM \cite{giannone2022few} is a few-shot sampling framework for conditional diffusion models. FSDM can adapt to different sampling processes and have a good sampling performance in a few steps by utilizing latent space of image transformers. Sehwag et al. \cite{sehwag2022generating}
guide the sampling process towards low-density regions and generated novel high fidelity samples from there. Also, Self-Attention Guidance \cite{hong2023improving} blurs the regions in the image which the model give attention and guide it accordingly with the residual information. 

Iterative Latent Variable Refinement (ILVR) \cite{choi2021ilvr} guides the generative process with a given reference image. They also found a solution to semantic correspondance problem by providing texts with a low-resolution image. Denoising Diffusion Restoration Models (DDRM) \cite{kawar2022denoising} also proposes a posterior sampling method which improves the quality and runtime in super-resolution, deblurring, inpainting, and colorization tasks. 

\subsubsection{Knowledge distillation}
Knowledge distillation \cite{hinton2015distilling} is an emerging method for obtaining efficient small-scale networks by transferring "knowledge" to simple student models from complex teacher models with high learning capacity \cite{gou2021knowledge}. Thus, student models have advantages as they are compressed and accelerated models. Denoising student \cite{luhman2021knowledge} distilled the multi-step denoising process into a single step, and achieved a similar sampling speed to other single-step generative models (GAN, VAE). In addition, this model does not have hyperparameters such as noising plan and step size. 

Salimans et al. has applied knowledge distillation to gradually distill information from one sampling model to another in Progressive Distillation \cite{salimans2022progressive}. At each distillation step, the student models are re-weighted by the teacher models before training, thus they are able to produce examples as close as the teacher models in one step. As a result, student models halve the sampling steps in each distillation process. They achieved competitive sampling successes in just four steps with the same training objective as DDPMs. 

Meng et al. \cite{meng2023distillation} focused on the computational cost of the classifier-free guided diffusion models such as Stable Diffusion \cite{rombach2022high}, DALL-E 2\cite{ramesh2022hierarchical} and Imagen \cite{saharia2022photorealistic} since they require evaluating a class-conditional model and an unconditional model over and over. They first match the output of the conditional and unconditional models, and then progressively distill information to another model that needs fewer sampling steps.

Berthelot et al. introduced Transitive Closure Time-distillation (TRACT) \cite{berthelot2023tract}, which extends binary time distillation and improved the speed of calculating FID by up to 2.4 times faster. Sun et al. distill feature distribution of the teacher model into the student in Classifier-based Feature Distillation (CFD) \cite{sun2023accelerating}. Also, Poole et al.  proposed score distillation sampling (SDS) \cite{poole2022dreamfusion} method that distill a text-to-image diffusion model to 3D neural radius fields (NeRF)\cite{mildenhall2020nerf}.

Many existing distillation techniques causes computation cost while generating synthetic data from the teacher model or need expensive online learning. BOOT \cite{gu2023boot} is an efficient data-free distillation algorithm that predicts the output of a teacher model in the given time steps.

Consistency model \cite{song2023consistency} supports directly mapping noise into data with one-step generation by distilling pre-trained diffusion models. It also allows to extend the sampling steps if one can afford computation cost to obtain more sample quality. Also Consistency Trajectory Models \cite{kim2023consistency} generalized Consistency models and Score-based models and propose a new sampling scheme involving long jumps in the ODE solution.

\section{Evaluation metrics \& Benchmark datasets}

\subsection{Evaluation metrics}
In this section, the most frequently used evaluation metrics will be discussed in order to evaluate the quality and  diversity of the samples produced and to compare the models with each other. 

\subsubsection{Inception Score (IS)}
Inception Score \cite{salimans2016improved}, is calcuted by using an Inception v3 network \cite{szegedy2016rethinking} pre-trained with ImageNet dataset \cite{deng2009imagenet}. This calculation can be divided into two parts as diversity measurement and quality measurement. A generated sample is considered to have the high resolution by how close to the corresponding class in the ImageNet dataset. For this reason, the similarity between the class images and the sample is calculated for quality measurement. On the other hand the diversity measurement, is calculated according to the class entropy of the produced samples. Greater entropy means the samples are more diverse. For the calculation of the Inception Score, which indicates a higher quality and more diverse sample production, the KL-divergence is  applied as in Equation \ref{eqn:inception_score}. Here, $p_{gen}$ is the generated distribution, and $p_{dis}$ is the distribution that the model knows. 

\begin{equation}
IS := exp\Big(E_{x\sim p_{gen}}\Big[D_{KL}\Big(p_{dis}(.|x) || \int p_{dis}(.|x)p_{gen}(x)dx \Big)\Big]\Big)
    \label{eqn:inception_score}
\end{equation}

\subsubsection{Frechet Inception Distance (FID)}
Although the Inception Score is a logical evaluation technique, it is based on a data set with 1000 classes and a network containing randomness trained with this data. The Frechet Inception Distance(FID)\cite{heusel2017gans} has been proposed to solve the bias of predetermined reference classes. In FID, the distance between the actual data distribution and the generated samples is calculated using mean and covariance. In other words, the latest  classification layer of the Inception v3 network is not used. The Frechet Inception  Distance is shown in Equation \ref{eqn:fid}. Here, $\mu_g$ and $\Sigma_g$ show the mean and covariance of the generated distribution, $\mu_r$ and $\Sigma_r$ show the mean and covariance of the distribution known to model. 

\begin{equation}
FID := ||\mu_r-\mu_g||_2^2+Tr\Big(\Sigma_r+\Sigma_g-2(\Sigma_r\Sigma_g)^{\frac{1}{2}}\Big)
    \label{eqn:fid}
\end{equation}

\subsubsection{Negative Log-likelihood (NLL)}
Negative log-likelihood (NLL) is a common evaluation criterion that can be used for all types of data  distributions. In order to determine how good our predictions are, class probabilities of the correct labels are taken into account. The purpose of logarithm calculation is to reduce the calculation cost by converting multiplications to addition. Many studies with VAE use NLL as an evaluation criterion\cite{razavi2019generating}. Since diffusion models can also explicitly model the distribution of data,  NLL calculation is possible. Even some diffusion models, such as the Improved DDPM \cite{nichol2021improved}, accept NLL as a training objective. If the model is denoted by $p_\theta$, the negative log-likelihood calculation can be expressed as in Equation \ref{eqn:nll}. 

\begin{equation}
NLL := E[-logp_\theta(x)]
    \label{eqn:nll}
\end{equation}

\subsection{Benchmark datasets}
In this section, we give the results of quality tests (benchmarks) in the most frequently used data sets. The results are given as FID(lower-is-better), IS(higher-is-better), and NLL(lower-is-better).

\subsubsection{CIFAR-10}
The CIFAR-10 dataset (Canadian Institute for Advanced Study, 10 classes) is a subset of the Tiny Images\cite{krizhevsky2009learning} dataset and consists of 60000 32x32 pixel color images. There are 6000 images total in each class, including 5000 training and 1000 test images. Table \ref{table:cifar-10} shows the results sorted by FID on the unconditional image generation task of CIFAR-10. EDM \cite{karras2022elucidating} shows superior performance in CIFAR-10 task and most of the state-of-the-art models are based on this model. Also this table shows us sampling-based approaches are more powerful than other methodologies. Space projection methods stand out by the performance among training-based approaches. 

\begin{center}
\begin{longtable}{| c | c c c |}

\hline \multicolumn{1}{|c|}{\textbf{Model}} & \multicolumn{1}{c}{\textbf{FID}} & \multicolumn{1}{c}{\textbf{IS}} & \multicolumn{1}{c|}{\textbf{NLL}} \\ \hline 
\endfirsthead

\hline
\multicolumn{4}{|l|}%
{ Continued from previous page} \\
\hline \multicolumn{1}{|c|}{\textbf{Model}} & \multicolumn{1}{c}{\textbf{FID}} & \multicolumn{1}{c}{\textbf{IS}} & \multicolumn{1}{c|}{\textbf{NLL}} \\
\hline 
\endhead

\hline \multicolumn{4}{|r|}{{Continued on next page}} \\ \hline
\endfoot

\endlastfoot

\hline
EDM-G++ \cite{kim2023refining} & 1.77 & - & 2.55 \\
EDM-ES \cite{ning2023elucidating} & 1.8 & - & - \\
CTM \cite{kim2023consistency} & 1.87 & - & 2.43 \\
STF \cite{xu2023stable} & 1.90 & - & - \\
LSGM-G++ (FID) \cite{kim2023refining} & 1.94 & - & 3.42 \\
EDM \cite{karras2022elucidating} & 1.97 & - & - \\
PSLD (ODE) \cite{pandey2023generative} & 2.10 & 9.93 & - \\
LSGM (FID) \cite{vahdat2021score} & 2.10 & - & 3.43 \\
ADM-ES \cite{ning2023elucidating} & 2.17 & - & - \\
Subspace Diffusion (NSCN++) \cite{jing2022subspace} & 2.17 & 9.94 & - \\
LSGM (balanced) \cite{vahdat2021score} & 2.17 & - & 2.95 \\
NCSN++ \cite{song2020score} & 2.20 & 9.89 & 3.45 \\
PSLD (SDE) \cite{pandey2023generative} & 2.21 & - & - \\
CLD-SGM (EM-QS) \cite{dockhorn2021score} & 2.23 & - & - \\
CLD-SGM (Prob. Flow) \cite{dockhorn2021score} & 2.25 & - & 3.31 \\
gDDIM \cite{zhang2022gddim} & 2.28 & - & - \\
INDM (FID) \cite{kim2022maximum} & 2.28 & - & 3.09 \\
Soft Truncation UNCSN++ (RVE) \cite{kim2022soft} & 2.33 & 10.11 & 3.04 \\
PFGM++ \cite{xu2023pfgm++} & 2.35 & 9.68 & 3.19 \\
BDDM \cite{lam2022bddm} & 2.38 & - & - \\
Subspace Diffusion (DDPM++) \cite{jing2022subspace} & 2.40 & 9.66 & - \\
SciRE-Solver-EDM(continuous) \cite{li2023sciresolver} & 2.40 & - & - \\
DDPM++ \cite{song2020score} & 2.41 & 9.68 & - \\
sub-VP-SDE \cite{song2020score} & 2.41 & 9.83 & 2.99 \\
Gotta Go Fast VP-deep \cite{jolicoeur2021gotta} & 2.44 & 9.61 & - \\
Soft Truncated DDPM++ (VP, FID) \cite{kim2022soft} & 2.47 & 9.78 & 2.91 \\
DEIS \cite{zhang2022fast} & 2.55 & - & - \\
Progressive Distillation \cite{salimans2022progressive} & 2.57 & - & - \\
DPM-Solver \cite{lu2022dpm} & 2.59 & - & - \\
DiffuseVAE-72M \cite{pandey2022diffusevae} & 2.62 & 9.75 & - \\
TDPM \cite{zheng2022truncated} & 2.83 & - & - \\
FastDPM \cite{kong2021fast} & 2.86 & - & - \\
Gotta Go Fast VE \cite{jolicoeur2021gotta} & 2.87 & 9.57 & - \\
iDDPM (FID) \cite{nichol2021improved} & 2.90 & - & 3.37 \\
Consistency Model \cite{song2023consistency} & 2.93 & 9.75 & - \\
Efficient Sampling \cite{watson2021learning} & 2.94 & - & - \\
SB-FBSDE \cite{chen2021likelihood} & 3.01 & - & 2.96 \\
PDM (VE, FID) \cite{kim2021maximum} & 3.04 & - & 3.36 \\
ES-DDPM \cite{lyu2022accelerating} & 3.11 & - & - \\
SciRE-Solver-EDM(discrete) \cite{li2023sciresolver} & 3.15 & - & - \\
DDPM \cite{ho2020denoising} & 3.17 & 9.46 & 3.72 \\
SN-DDIM \cite{bao2022estimating} & 3.22 & - & - \\
INDM (ST) \cite{kim2022maximum} & 3.25 & - & 3.01 \\
PNDM \cite{liu2022pseudo} & 3.26 & - & - \\
SN-DDPM \cite{bao2022estimating} & 3.31 & - & - \\
Calibrating DPM-Solver Discrete \cite{pang2023calibrating} & 3.31 & - & - \\
Analytic DDIM \cite{bao2022analytic} & 3.39 & - & - \\
NPR-DDIM \cite{bao2022estimating} & 3.42 & - & - \\
DPM-Solver Discrete \cite{lu2022dpm} & 3.45 & - & - \\
Soft Truncation DDPM++ (VP, NLL) \cite{kim2022soft} & 3.45 & 9.19 & 2.88 \\
Analog Bits \cite{chen2023analog} & 3.48 & - & - \\
NPR-DDPM \cite{bao2022estimating} & 3.57 & - & 3.79 \\
GENIE \cite{dockhorn2022genie} & 3.64 & - & - \\
DSM-EDS \cite{jolicoeur2020adversarial} & 3.65 & - & - \\
Diffusion Step \cite{franzese2022much} & 3.72 & - & 3.07 \\
Denoising Diffusion GAN \cite{xiao2021tackling} & 3.75 & 9.63 & - \\
TS-DDIM (quadratic) \cite{li2023alleviating} & 3.81 & - & - \\
Soft Diffusion \cite{daras2022soft} & 3.86 & - & - \\
ScoreFlow (VP, FID) \cite{song2021maximum} & 3.98 & - & 3.04 \\
WaveDiff \cite{phung2022wavelet} & 4.01 & - & - \\
DDIM \cite{song2020denoising} & 4.04 & - & - \\
GGDM \cite{watson2021learningiclr} & 4.25 & 9.19 & - \\
Analytic DDPM \cite{bao2022analytic} & 4.31 & - & 3.42 \\
Blackout Diffusion \cite{santos2023blackout} & 4.58 & 9.01 & - \\
INDM (NLL) \cite{kim2022maximum} & 4.79 & - & 2.97 \\
Dynamic Dual-Output \cite{benny2022dynamic} & 5.10 & - & - \\
Diff-Pruning \cite{fang2023structural} & 5.29 & - & - \\
ScoreFlow (deep, sub-VP, NLL) \cite{song2021maximum} & 5.40 & - & 2.81 \\
PDM (VP, NLL) \cite{kim2021maximum} & 6.84 & - & 2.94 \\
LSGM (NLL) \cite{vahdat2021score} & 6.89 & - & 2.87 \\
D3PM \cite{austin2021structured} & 7.34 & 8.56 & 3.44 \\
Denoising student \cite{luhman2021knowledge} & 9.36 & 8.36 & - \\
EBM-DRL \cite{gao2020learning} & 9.58 & 8.30 & - \\
NCSNv2 \cite{song2020improved} & 10.87 & 8.40 & - \\
iDDPM (NLL) \cite{nichol2021improved} & 11.47 & - & 2.94\\
DiffFlow \cite{zhang2021diffusion} & 14.14 & - & 3.04 \\
DAED \cite{deja2022analyzing} & 14.2 & 8.6 & - \\
NCSN \cite{song2019generative} & 25.32 & 8.87 & - \\
VDM \cite{kingma2021variational} & - & - & 2.49 \\
\hline
\caption{CIFAR-10 image generation} 
\label{table:cifar-10}
\end{longtable}
\end{center}

\subsubsection{CelebA}
The CelebFaces Attributes dataset\cite{liu2015deep} contains 202.599 facial images from 10.177 celebrities that are 178×218 pixel. Each of the images has 40 binary labels showing facial features such as hair color, gender and age. The 64x64 pixel version of the CelebA dataset is used more frequently as a benchmark in the image generation with diffusion models. Table \ref{table:celeba} shows the results sorted by FID. On the contrary of CIFAR-10 training-based approaches are shown superior performance in this task. 

\begin{table}
    \centering
\begin{threeparttable}[htbp]
    \begin{tabular}{| c | c c |}
\hline
\textbf{Model} & \textbf{FID} & \textbf{NLL} \\
\hline
DDPM-IP \cite{ning2023input} & 1.27 & - \\
STDDPM-G++ \cite{kim2023refining} & 1.34 & - \\
INDM (VP, FID) \cite{kim2022maximum} & 1.75 & 2.27 \\
Soft Diffusion(VE-SDE + Blur) \cite{daras2022soft} & 1.85 & - \\
Soft Truncated DDPM++ (VP, FID) \cite{kim2022soft} & 1.90 & 2.10 \\
Soft Truncated UNCSN++ (RVE) \cite{kim2022soft} & 1.92 & 1.97 \\
SciRE-Solver \cite{li2023sciresolver} & 2.02 & - \\
PDM (VP, FID) \cite{kim2021maximum} & 2.04 & 2.23 \\
Calibrating DPM-Solver \cite{pang2023calibrating} & 2.33 & - \\
PDM (VE, FID) \cite{kim2021maximum} & 2.50 & 2.00 \\
INDM (VE, FID) \cite{kim2022maximum} & 2.54 & 2.31 \\
ES-DDPM \cite{lyu2022accelerating} & 2.55 & - \\
DPM-Solver Discrete \cite{lu2022dpm} & 2.71 & - \\
PNDM \cite{liu2022pseudo} & 2.71 & \\
SN-DDIM \cite{bao2022estimating} & 2.85 & - \\
Soft Truncated DDPM++ (VP, NLL) \cite{kim2022soft} & 2.90 & 1.96 \\
Gamma Distribution DDIM \cite{nachmani2021non} & 2.92 & \\
INDM (VP, NLL) \cite{kim2022maximum} & 3.06 & 2.05 \\
Analytic DDIM \cite{bao2022analytic} & 3.13 & - \\
NPR-DDIM \cite{bao2022estimating} & 3.15 & - \\
Blackout Diffusion \cite{santos2023blackout} & 3.22 & - \\
DDPM \cite{ho2020denoising}\tnote{1} & 3.26 & - \\
DDIM \cite{song2020denoising} & 3.51 & \\
Mixture Gaussian DDIM \cite{nachmani2021non} & 3.71 & \\
NCSN++ \cite{song2020score}\tnote{2} & 3.95 & 2.39 \\
DiffuseVAE \cite{pandey2022diffusevae} & 3.97 & - \\
Dynamic Dual-Output \cite{benny2022dynamic} & 4.07 & - \\
Gamma Distribution DDPM \cite{nachmani2021non} & 4.09 & \\
TS-DDIM (quadratic) \cite{li2023alleviating} & 4.18 & - \\
SN-DDPM \cite{bao2022estimating} & 4.42 & - \\
Analytic DDPM \cite{bao2022analytic} & 5.21 & 2.66 \\
NPR-DDPM \cite{bao2022estimating} & 5.33 & 2.65 \\
Mixture Gaussian DDPM \cite{nachmani2021non} & 5.57 & \\
EBM-DRL \cite{gao2020learning} & 5.98 & - \\
Diff-Pruning \cite{fang2023structural} & 6.24 & - \\
FastDPM \cite{kong2021fast} & 7.85 & - \\
NCSNv2 \cite{song2020improved} & 10.23 & - \\
DAED \cite{deja2022analyzing} & 15.1 & - \\
NCSN \cite{song2019generative}\tnote{3} & 25.30 & - \\
\hline
    \end{tabular}
    \begin{tablenotes}
    \item[1] Reported by Song et al. \cite{song2020denoising}   
    \item[2] Reported by Kim et al. \cite{kim2022soft}
    \item[3] Reported by Song et al. \cite{song2020improved}
  \end{tablenotes}
    \caption{CelebA(64X64) image generation}
    \label{table:celeba}
\end{threeparttable}
\end{table}

\subsubsection{ImageNet}
The ImageNet dataset \cite{deng2009imagenet} contains 14.197.122 annotated images according to the WordNet hierarchy. Since 2010, it has been used in the "ImageNet Large-Scale Visual Recognition Competition (ILSVRC)", as a criterion for image classification and object recognition. It contains binary labels that show presence or absence of a class of objects in the image contain for image-level labeling task. It also contain a tight bounding box and a class label around an object instance in the image for an object-level labeling task. Table \ref{table:ImageNet} shows the benchmarks sorted by FID on the image generation task of 64x64 pixel ImageNet dataset. The models are divided into two groups according to the class conditional and unconditional generation tasks. In the context of conditional generation, most of the best approaches are training-based. Sampling-based approaches are competitive in this task and also shown better performance on unconditional generation task.

\begin{table}[htbp]
    \centering
    \begin{tabular}{| c | c | c c c |}
\hline
\textbf{Task} & \textbf{Model} & \textbf{FID} & \textbf{IS} & \textbf{NLL} \\
\hline
\multirow{11}{*}{Conditional} 
& VDM++ \cite{kingma2023understanding} & 1.43 & 64.6 & - \\
& CDM \cite{ho2022cascaded} & 1.48 & 67.95 & - \\
& CTM \cite{kim2023consistency} & 1.90 & - & 63.90 \\
& ES-DDPM \cite{lyu2022accelerating} & 2.07 & 55.29 & - \\
& ADM \cite{dhariwal2021diffusion} & 2.07 & - & - \\
& iDDPM \cite{nichol2021improved} & 2.92 & - & - \\
& Consistency Model \cite{song2023consistency} & 4.70 & - & - \\
& Analog Bits \cite{chen2023analog} & 4.84 & - & - \\
& Self-guided DM \cite{hu2023selfguided} & 12.1 & 23.1 & - \\
& BOOT \cite{gu2023boot} & 16.3 & - & - \\
& Classifier-free guidance \cite{ho2022classifier} & 26.22 & 260.2 & - \\
\hline
\multirow{13}{*}{Unconditional}
& Analytic-DDPM \cite{bao2022analytic} & 16.14 & - & 3.61 \\
& SN-DDPM \cite{bao2022estimating} & 16.22 & - & -\\
& NPR-DDPM \cite{bao2022estimating} & 16.32 & - & 3.71 \\
& TS-DDIM (quadratic) \cite{li2023alleviating} & 17.20 & - & - \\
& SN-DDIM \cite{bao2022estimating} & 17.23 & - & -\\
& NPR-DDIM \cite{bao2022estimating} & 17.30 & - & - \\
& Analytic-DDIM \cite{bao2022analytic} & 17.44 & - & - \\ 
& DPM-Solver Discrete \cite{lu2022dpm} & 17.47 & - & - \\
& DDIM \cite{song2020denoising} & 17.73 & - & - \\
& GGDM \cite{watson2021learningiclr} & 18.4 & 18.12 & - \\
& iDDPM \cite{nichol2021improved} & 19.2 & - & 3.57 \\
& VDM \cite{kingma2021variational} & - & - & 3.40 \\
& Efficient Sampling \cite{watson2021learning} & -  & - & 3.55 \\
\hline
    \end{tabular}
    \caption{ImageNet(64x64) image generation}
    \label{table:ImageNet}
\end{table}

\section{Conclusion and Future Directions}
This study provides a comprehensive and up-to-date review of the generative diffusion models. We start by giving a brief information about generative models, we discussed the diffusion models, why we need them and the advantages/disadvantages over other generative models. We give an introduction to diffusion models with three main studies: DDPMs, NCSNs, and Score SDEs. Next, we divide the studies to improve diffusion models into 2: improvements in the training process and improvements on the training-free sampling. Finally, we give evaluation criteria and benchmark results of the mentioned studies. 

Research about diffusion models are developing rapidly, both theoretically and practically. There are some following paths for future research on diffusion models.

It is assumed that forward process of diffusion models completely erases the whole information about the data and results in a random distribution, but this may not always be the case. In reality, the complete conversion of information to noise cannot be accomplished in a finite time.

It is important to understand why and when diffusion models are effective while setting a forward progress path. Understanding the key features that distinguish diffusion models from other types of generative models will help to explain why they are more successful in the certain tasks. In this manner theoretical guidance is also required to systematically select and determine the various hyperparameters of diffusion models. For example, determining when to stop the noise process is an important factor to balance between sampling speed and sample quality.

Most diffusion models take ELBO as training objective. However, the issue of optimizing ELBO and NLL simultaneously is controversial. More advanced analytical approaches that link log-likelihood optimization to existing variables or use new probability-consistent training objectives rather than ELBO can be improve the performance.

Data distributions are not covariant with likelihood matching so there is a requirement for some evaluation criteria that more accurately and comprehensively determine sample diversity and the production capability of the model. Stein et al. \cite{stein2023exposing} discussed current evaluation metrics and assert that these metrics are not detecting memorization.

About the network design there are hybrid models with vision transformers(ViT), U-ViT \cite{bao2023worth}, GenViT and HybViT \cite{yang2022vit} provide long skip connections which is crucial for learning large scale multimodal datasets with diffusion models. There can be also an improvement by using different network structures.

The researchers have not access to uncorrupted samples or it is expensive to have them in some application areas. Daras et al. \cite{daras2023ambient} introduced additional measurement
distortion in the diffusion process thus  the model generate samples using only highly-corrupted samples. By the further developments, diffusion models can help the researchers in these fields.

Beside the application success, diffusion models have also a powerful background to investigate. Denoising Diffusion Samplers (DDS) \cite{vargas2023denoising} provide theoretical guarantees by approximating the corresponding time-reversal with Monte Carlo sampling. There are also some studies \cite{oko2023diffusion, chen2023sampling,de2022riemannian} provide theoretical guarantees for well-known function spaces with diffusion modeling. Also McAllester \cite{mcallester2023mathematics} presents mathematical understanding of diffusion models. All of these approaches have contributed to the algorithm of the diffusion models. This new generative model family is developing rapidly and there would be so many improvements which boost them in several fields.



 \bibliographystyle{plainnat} 
 \bibliography{cas-refs}





\end{document}